%% file: main.tex
\documentclass{article}

\input{sections/00-preamble}      %

\begin{document}

\input{sections/01-frontmatter}   %
\input{sections/02-introduction}  %
\input{sections/03-methodology}   %
\input{sections/04-results}       %
\input{sections/05-discussion}    %
\input{sections/06-conclusion}    %

\newpage
\vskip 0.2in
\bibliography{references}

\end{document}

%% file: sections/00-preamble.tex
\usepackage{arxiv}          %

\usepackage[utf8]{inputenc}
\usepackage[T1]{fontenc}
\usepackage{microtype}
\usepackage{amsmath}
\usepackage{amssymb}
\usepackage{amsfonts}       %
\usepackage{graphicx}
\usepackage{empheq}
\usepackage{xcolor}
\usepackage{makecell}
\usepackage{neuralnetwork}
\usepackage{csquotes}
\usepackage{booktabs}       %
\usepackage{multirow}       %
\usepackage{siunitx}
\usepackage{float}
\usepackage{url}
\usepackage[hidelinks]{hyperref}  %

\usepackage[numbers,sort]{natbib}
\DeclareMathOperator{\enc}{enc}
\DeclareMathOperator{\dec}{dec}

\renewcommand{\headeright}{A Preprint}
\renewcommand{\undertitle}{A Preprint}
\renewcommand{\shorttitle}{Partially Linear Autoencoders}

%% file: sections/01-frontmatter.tex
\title{Partially Linear Autoencoders for Manifold Learning\\
and Dimensionality Reduction}

\author{
  Louen Pottier\\
  CEA List, Université Paris-Saclay\\
  Palaiseau, F-91120, France\\
  \texttt{louen.pottier@cea.fr}
  \And
  Louis Lesueur\\
  LMS, École Polytechnique, Institut Polytechnique de Paris\\
  Route de Saclay, Palaiseau, F-91128, France\\
  \texttt{louis.lesueur.lms@polytechnique.edu}
  \And
  Anders Thorin\\
  CEA List, Université Paris-Saclay\\
  Palaiseau, F-91120, France\\
  \texttt{anders.thorin@cea.fr}
}

\date{\today}

\hypersetup{
  pdftitle={Partially Linear Autoencoders for Manifold Learning and
            Dimensionality Reduction},
  pdfauthor={Louen Pottier, Louis Lesueur, Anders Thorin},
  pdfkeywords={autoencoders, dimensionality reduction, manifold learning},
}

\maketitle

\newcommand\rka[1]{\textcolor{blue}{#1}}
\newcommand\rkl[1]{\textcolor{red}{#1}}

\sisetup{
  detect-weight = true,        %
  round-precision = 3,         %
  table-number-alignment = center
}

\begin{abstract}%
Autoencoders are widely used for nonlinear dimensionality reduction and manifold learning. While most common implementations rely on both nonlinear encoders and decoders, we investigate the specific role of the encoder and the extent to which it can be constrained to be linear without reducing accuracy. We conduct a comparative study on four autoencoder architectures: standard fully nonlinear autoencoders (AE), linear-encoder autoencoders (Lenc-AE), linear-decoder autoencoders (Ldec-AE), and fully linear autoencoders (LAE), evaluated on synthetic manifolds, computational mechanics data sets, and real-world image data sets including MNIST.
We demonstrate that imposing a linear encoder preserves most of the representational capacity of the autoencoder, provided the decoder remains nonlinear. In particular, Lenc-AE consistently outperforms both Ldec-AE and LAE, and achieves reconstruction quality comparable to fully nonlinear AE, while offering advantages in terms of parsimony and interpretability of the latent representation. These results suggest that the nonlinear decoder is the critical component for manifold learning, rather than the encoder.
A geometric interpretation of this finding is developed, which identifies the precise conditions under which a linear encoder is sufficient, and the specific manifold configurations that expose its limitations.
\end{abstract}%

\keywords{autoencoders \and dimensionality reduction \and manifold learning}

%% file: sections/02-introduction.tex
\section{Introduction}
\label{sec:introduction}

Dimensionality reduction is a cornerstone of modern data analysis, enabling the extraction of meaningful patterns from high-dimensional data sets. Consider a data set \( \mathcal{X} = \{\mathbf{x}_i\}_{i=1}^n \subset \mathbb{R}^D \), where each \( \mathbf{x}_i \) represents a sample in a high-dimensional space (an image for example). Traditional approaches assume that such data lies in a low-dimensional linear subspace of dimension \( d \ll D \). This assumption underpins methods like Principal Component Analysis (PCA) \citep{jolliffe2002principal}, which seeks an orthogonal projection maximizing explained variance.%

However, this linearity assumption can be overly restrictive for complex data across diverse domains such as natural images, stellar spectra, biomedical signals, or genomic data, where significant variations (textures, rotations, temporal dynamics, or high-dimensional correlations) cannot be easily captured by a single linear subspace, highlighting the need for more expressive nonlinear dimensionality reduction methods.
Manifold learning methods~\citep{ManifoldLearning}, including Isomap~\citep{isomap}, Locally Linear Embedding (LLE)~\citep{LLE}, t-SNE~\citep{JMLR:v9:vandermaaten08a} and UMAP~\citep{mcinnes2020umapuniformmanifoldapproximation}, address this challenge by relying on the manifold hypothesis: the assumption that high-dimensional data lies on or near a low-dimensional $p$-submanifold $\mathcal{M} \subset \mathbb{R}^D$. These approaches exploit local neighborhoods to construct embeddings that preserve the intrinsic geometry of $\mathcal{M}$. For example, Isomap approximates geodesic distances on the manifold, while t-SNE and UMAP focus on preserving local neighborhood relationships through probabilistic and topological frameworks, respectively. Such methods reveal complex structures—like clusters, nonlinear trends, or multi-scale organizations—that linear techniques such as PCA fail to capture. A canonical example is the Swiss~roll~\citep{isomap,LLE,ManifoldLearning}, a two-dimensional manifold embedded in $\mathbb{R}^3$. While its intrinsic dimension is $p=2$, its coiled structure renders linear projections (e.g., PCA) ineffective for dimension reduction. Nonlinear methods, however, successfully ``unroll'' the manifold. 
Classical manifold learning methods generally provide only an implicit embedding of the training samples and do not define an explicit, parametric mapping from the high-dimensional space to the reduced space. As a result, extending these embeddings to out-of-sample data, or using them for model reduction or inverse problems, remains challenging.
 
\subsection{Autoencoders}
 
In this context, nonlinear autoencoders (AEs) \citep{Goodfellow-et-al-2016,shi2021discussionvaliditymanifoldlearning} have emerged as a powerful alternative to both linear methods and classical manifold learning approaches, capable of capturing nonlinear relationships through deep neural architectures while providing an explicit, parametric mapping that generalizes to out-of-sample data. An autoencoder consists of two learned functions: an encoder \( \enc: \mathbb{R}^D \rightarrow \mathbb{R}^d \) that projects data into a low-dimensional latent space, and a decoder \( \dec: \mathbb{R}^d \rightarrow \mathbb{R}^D \) that reconstructs the original data from its latent representation. Training minimizes a reconstruction loss, typically \( \frac{1}{N} \sum_{i=1}^N \|\dec(\enc(\mathbf{x}_i)) - \mathbf{x}_i\|^2 \).

A geometric perspective on how autoencoders learn manifolds is provided~\citep{lee2023geometricperspectiveautoencoders}, highlighting the role of nonlinear mappings in capturing complex data structures. While this approach overcomes the limitations of linear methods, it introduces new challenges:
\begin{itemize}
    \item Choosing \( d \): The latent dimension must balance underfitting and compression efficiency. The Whitney embedding theorem \citep{whitney} provides a theoretical bound: a \( p \)-dimensional differential manifold can be embedded in \( \mathbb{R}^{2p} \) without self-intersections. If the manifold is homeomorphic to \( \mathbb{R}^d \), \( d = p \) is sufficient. However, \( p \) is often unknown and must be estimated.
    \item Structuring the latent space \( \mathcal{N} = \enc(\mathcal{X}) \): Without explicit constraints, the learned latent representation \( \mathcal{N} \) may lack exploitable geometric properties. For applications such as model coupling or solving differential equations in reduced coordinates, preserving the manifold structure in latent space is crucial. Several approaches~\citep{DBLP:journals/corr/abs-2006-09289, umapae, Braunsmann_2024} incorporate distance-preserving regularization terms in the training objective, encouraging the encoder to preserve local pairwise distances, thereby approximating an isometric mapping in the latent space. Other approaches focus on class separation in supervised settings. For instance, the reconstruction loss has been modified to encourage latent representations of the same class to cluster together, requiring labeled data~\citep{UndercompleteAutoencoder}. %
\end{itemize}

\begin{figure}[ht!]
    \centering
    \includegraphics[width=\textwidth]{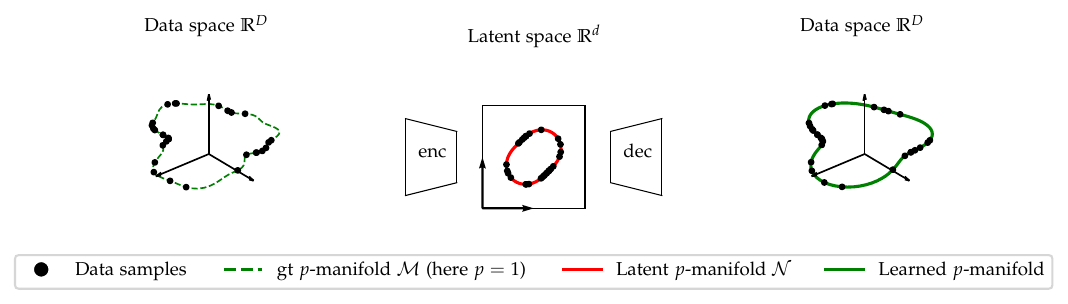}
    \caption{Schematic representation of an autoencoder learning a nonlinear $p$-manifold (here $p=1$). The encoder $\enc$ projects data samples $\mathcal{X} \subset \mathbb{R}^D$ (here $D=3$) from the ground truth $p$-manifold $\mathcal{M}$ (green dashed curve) to a low-dimensional latent space $\mathcal{N} \subset \mathbb{R}^d$ (here, $d=2$), which preserves the manifold structure. The decoder $\dec$ reconstructs the manifold (green curve) from latent coordinates.}
    \label{fig:manifold}
\end{figure}

When the training data set is well-sampled on a known manifold and the autoencoder architecture is appropriately designed—with a latent dimension $d$ consistent with Whitney's embedding theorem and sufficient expressivity—the latent space faithfully preserves the manifold's structure, as illustrated in Figure~\ref{fig:manifold}.

\subsection{Asymmetric Autoencoder Architectures and the Role of Encoder vs. Decoder Complexity}
Most standard autoencoder implementations employ symmetric architectures: encoders and decoders with identical numbers of layers, comparable hidden dimensions, and mirrored activation patterns (exemplified by U-Net-style architectures)~\citep{Hinton2006, Goodfellow-et-al-2016}. This design choice reflects a natural intuition—that encoding and decoding should be inverse operations—but it is primarily driven by practical convenience rather than fundamental necessity. The high computational cost of exhaustive hyperparameter search, combined with the multiplicity of architectural choices (number of layers, hidden dimensions, activation functions), has made symmetric designs an attractive heuristic that has become the default in many works.

However, the encoder and decoder play fundamentally different roles in manifold learning. The encoder's role is to produce a compact, injective projection of the data manifold onto latent space; the decoder must reconstruct the manifold's full geometric complexity from that projection. These are asymmetric tasks by nature, and there is no a priori reason they should require equal computational resources or parametric complexity. 
Accordingly, the optimal allocation of capacity between encoder and decoder is task-dependent. When the input space lacks intrinsic geometric structure—as in natural image data sets, where semantic similarity must be learned rather than inherited from the data geometry—the encoder bears the greater burden: it must discover and impose a meaningful organization of the latent space \citep{zhang2022surveymaskedautoencoderselfsupervised}. Conversely, when the input data already lies on a structured space—where geometric proximity reflects similarity—the encoding task is comparatively straightforward, and the complexity shifts to the decoder, which must learn to map a low-dimensional latent representation back to a high-dimensional output space.
The appropriate architecture is therefore not a matter of symmetry, but of matching component capacity to component function.

\subsection{Motivation and Advantages of Partial Linearity}

While the majority of autoencoder implementations employ both nonlinear encoders and nonlinear decoders, many applications impose or benefit from constraints on one of these components. This motivates a systematic investigation of the four fundamental architectures that result from independently choosing linear or nonlinear encoders and decoders, as summarized in Table~\ref{tab:architectures}.

\begin{table}[h]
\centering
\begin{tabular}{|l|c|c|}
\hline
\diaghead{\theadfont Diag ColumnmnH}{\textbf{Encoder}}{\textbf{Decoder}} &
{Linear} & {Nonlinear (MLP)} \\
\hline
Linear & LAE & Lenc-AE \\
\hline
Nonlinear (MLP) & Ldec-AE & AE \\
\hline
\end{tabular}
\caption{Taxonomy of autoencoder architectures compared in this work, based on encoder and decoder linearity.}
\label{tab:architectures}
\end{table}

It is worth noting that the fully linear autoencoder (LAE) is mathematically equivalent to Principal Component Analysis (PCA)~\citep{plaut2018principalsubspacesprincipalcomponents}: minimizing reconstruction loss with both linear encoder and decoder recovers the principal subspace through gradient-based optimization. In practice, one would simply perform PCA rather than train an LAE. The three remaining architectures—Lenc-AE, Ldec-AE, and AE—offer genuinely distinct approaches to nonlinear dimensionality reduction.

Partially linear architectures occupy an intermediate position between the simplicity of PCA (LAE) and the full expressivity of nonlinear autoencoders (AE). This intermediate regime not only offers a compelling trade-off between representational capacity and practical advantages—such as interpretability, computational efficiency, or alignment with physical constraints—but also arises naturally across diverse domains. Indeed, in fields where data-driven dimensionality reduction must balance expressivity with these practical considerations, partially linear autoencoders have emerged as a preferred approach. We review key examples from several domains to contextualize our comparative study.

\subsubsection{General Applications}
Partially linear autoencoders have been widely studied in machine learning. Reduced-complexity designs with linear encoders for efficient signal recovery were explored~\citep{ZOCCO2022104663}, and asymmetric Lenc-AE architectures were proposed to reduce computational cost in edge computing, leveraging nonlinear decoders for offline reconstruction~\citep{GILBERT2024103412}.  A sparse recovery autoencoder with linear encoder projections was introduced~\citep{SparseRecovery}, and the effectiveness of Ldec-AE in subspace clustering by exploiting block-diagonal latent structures and closed-form clustering optimization was demonstrated~\citep{9263249}.

\subsubsection{Model Reduction in Structural Mechanics}
In computational mechanics, classical model reduction techniques align with autoencoder (AE) architectures. Proper Orthogonal Decomposition (POD)~\citep{POD}—equivalent to PCA on temporal snapshots—corresponds to a linear autoencoder (LAE), offering data-driven dimensionality reduction.
Component mode synthesis methods such as \citet{guyan} and Craig–Bampton~\citep{craigbrampton} resemble partially linear AEs: they encode boundary/interface degrees of freedom via physics-informed linear transformations, while decoding internal states. Their linearity at the encoder level is physically necessary for substructure coupling compatibility. Extending these methods with nonlinear decoders (e.g., neural networks) could yield Lenc-AE architectures, though this remains unexplored for capturing nonlinear effects (e.g., large deformations) while preserving interpretability.
Recent advances combine POD with neural networks to model dynamical systems. POD was paired with neural ODEs for reduced-order dynamics~\citep{lee2019modelreductiondynamicalsystems}, implementing Ldec-AE or AE architectures. POD-based decoders were used with neural latent dynamics~\citep{dolz2025fullydiscreteanalysisgalerkin}, while Latent Energy-Based Neural Networks (LEBNN)—fully nonlinear AEs preserving non-convex energy landscapes—have been proposed~\citep{LEBNN}. Latent Lagrangian Neural Networks were developed~\citep{friedl2025riemannianframeworklearningreducedorder}, embedding geometric constraints via Riemannian frameworks.
A key challenge is ensuring latent space regularity for stable ODE/PDE integration. Linear encoders (Lenc-AE) inherently produce structured latent spaces, preserving affine properties (e.g., trajectory linearity, distance ratios) without explicit regularization. This contrasts with nonlinear encoders, which may learn distorted representations complicating numerical integration. Regularization strategies (e.g., distance-preserving losses~\citep{DBLP:journals/corr/abs-2006-09289, Braunsmann_2024}, UMAP-inspired losses~\citep{umapae}) further enforce geometric structure, but Lenc-AE’s intrinsic regularity makes it particularly suitable for reduced-order dynamical systems.

\subsubsection{Dynamical Systems and Koopman Operator Theory} 
Koopman operator methods for analyzing nonlinear dynamical systems provide another context where partially linear architectures emerge naturally. Dynamic Mode Decomposition (DMD)~\citep{SCHMID_2010} approximates linear dynamics directly in the state space, without dimensionality reduction. Extended DMD (EDMD)~\citep{Williams_2015} introduces a dictionary of nonlinear observable functions $\Psi(x)$ that are evaluated on the data (forming a nonlinear encoding), evolved with linear dynamics in feature space, and used to reconstruct the state via linear combinations with Koopman modes—a structure equivalent to Ldec-AE. Kernel DMD (KDMD)~\citep{williams2015kernelbasedapproachdatadrivenkoopman} extends this framework to implicit, potentially infinite-dimensional feature spaces using kernel functions. 
Koopman-based modeling of dynamical systems, replacing EDMD's linear decoder with a nonlinear neural network (their Linearly-Recurrent Autoencoder Network, or LRAN) was shown to enable accurate state reconstruction from fewer latent features~\citep{otto2019linearlyrecurrentautoencodernetworkslearning}, as the full state no longer needs to lie within a finite-dimensional Koopman-invariant subspace. This progression from EDMD (Ldec-AE) to LRAN (AE with linear recurrent dynamics) demonstrates the advantage of concentrating nonlinearity in the decoder when learning representations of dynamical systems. More recent work on dictionary learning for EDMD~\citep{Li_2017} uses neural networks to learn adaptive dictionaries, further improving the expressivity of the encoding while maintaining linear reconstruction.

\subsubsection{Other Domain-specific Applications}
In certain application domains, the choice between linear and nonlinear components is dictated by interpretability constraints.  For example, in hyperspectral unmixing, observed pixel spectra are modeled as linear combinations of pure material spectra, known as endmembers. The goal of unmixing is to estimate the proportional contributions—abundances—of these endmembers for each pixel. By fixing the weights of a linear decoder to the known endmember spectra, the autoencoder directly encodes the physical linearity of the mixing process, ensuring that the reconstructed spectra adhere to the constraints of the linear mixing model \citep{8984691,9096565}. Similarly, in medical imaging, a linear decoder allows for greater transparency in how latent features combine to form reconstructions \citep{VIVEK2023104536}. Clinicians can more easily interpret the contributions of individual features to the final output, which is critical for diagnostic decision-making and trust in automated systems.

\subsubsection{Summary of the Review}
As highlighted in the previous state-of-the-art review, linear encoder or linear decoder architectures offer several notable advantages: \begin{itemize}
\item Interpretability: The two architectures offer complementary forms of interpretability. A linear encoder allows direct analysis of which input features contribute to each latent dimension—the projection weights are immediately readable as a basis decomposition of the input space. A linear decoder offers a different transparency: the reconstruction is an explicit linear combination of basis vectors, making it straightforward to inspect which latent directions govern which output patterns. In both cases, linearity in one component provides an analytical handle that fully nonlinear architectures lack.
\item Computational efficiency: The linear component (whether encoder or decoder) reduces to a single matrix-vector product, enabling real-time execution on resource-constrained hardware. This is particularly relevant in asymmetric deployment scenarios: a lightweight linear encoder can perform on-device compression while reconstruction is handled offline, or conversely, a linear decoder enables fast on-device reconstruction from latent codes computed elsewhere.
\item Domain-specific structural constraints: Many physical and engineering applications impose explicit constraints on the form of the encoder or decoder. A linear encoder ensures that interface coordinates appear linearly in the reduced representation, as required by structural coupling methods (Craig–Bampton, Guyan reduction). A linear decoder constrains the reconstruction to a fixed linear span, which is consistent with Koopman operator theory and projection-based reduced-order models. In both cases, the linearity constraint is not a limitation but an alignment with the mathematical structure of the target application.
\end{itemize}
The two architectures differ, however, in how they handle geometric regularity of the latent space. A linear encoder preserves the geometric structure of the input space—distances, parallelism, and affine relationships—so the latent space is regular precisely when the input manifold is. This is a natural fit for physical simulation, where the solution space inherits regularity from the underlying PDE. A linear decoder, by contrast, forces the encoder to find a latent representation from which the output can be linearly reconstructed—effectively learning coordinates adapted to the data geometry rather than inheriting them from the input space.

\subsubsection{Motivation and Contributions}

The observations made above lead us to question to what extent a linear encoder combined with a nonlinear decoder (Lenc-AE) can match the reconstruction quality of fully nonlinear autoencoders (AEs). Additionally, they prompt a comparison between Lenc-AE and Ldec-AE to determine which component, encoder or decoder, plays the more critical role in capturing nonlinear manifold structures. It also raises the question of under which geometric conditions a linear encoder becomes a limiting factor.

To address these problems, we first introduce architectures and experimental protocols in Section~\ref{sec:methodology}. Next, Section~\ref{sec:results} presents comparative results, highlighting the performance of Lenc-AE. Finally, Section~\ref{sec:discussion} discusses the broader implications of these findings, particularly for model reduction in mechanics, and outlines directions for future research.

%% file: sections/03-methodology.tex
\section{Methodology}
\label{sec:methodology}

To investigate the questions raised in the introduction, we conduct a comparative study on three types of data, covering eight test cases in total:
\begin{itemize}
    \item Synthetic manifolds (helix, knot, Swiss roll, S-curve), where the intrinsic dimension and geometric structure are known, enabling fine-grained analysis of reconstruction capabilities.
    \item Two computational mechanics examples: reconstruction of displacement fields of a cantilever beam and a paperclip, where the latent dimension is tied to physical parameters (e.g., boundary conditions). 
    \item Computer vision examples: the MNIST and Office data sets (see Section~\ref{sec:office} for a detailed description of the latter), to assess scalability on high-dimensional real-world data.
\end{itemize}

A summary of the characteristics of all these data sets is given in Table~\ref{tab:characteristics}.
For the Beam test case, the intrinsic dimension is known ($p=2$) since it corresponds to the number of physical parameters used to generate the displacement fields (see the data set of~\citet{LEBNN}). The Office data set is a collection of 477 images from a single video of an office, taken by rotating the camera by 360 degrees around a vertical axis while keeping it horizontal. This set of images theoretically forms a closed curve in the abstract set of $(3 \times 128 \times 128)$-shaped images and therefore has an intrinsic dimension $p=1$.

In contrast, for the Paperclip (see the data set of~\citet{HQJMYA_2025}) and MNIST data sets, the intrinsic dimension of the problem is unknown. It can only be estimated from the minimal latent dimension required for the autoencoder to correctly reconstruct the input data.  For MNIST, several studies have attempted to estimate its intrinsic dimension, typically reporting values between 10 and 15 depending on the methodology~\citep{pope2021intrinsicdimensionimagesimpact, jing2020implicitrankminimizingautoencoder, zeng2023autoencodersdiscoveringmanifolddimension}.
\begin{table}[h]
\centering
\begin{tabular}{lccc}
\hline
data set & original dimension $D$ & dimension as manifold $p$ & latent space dimension $d$ \\ \hline 
Helix & 3 & 1 & \{1,2\}\\
Knot & 3 & 1 & \{1,2,3\}\\
S-curve & 3 & 2 & 2\\
Swiss roll & 3 & 2 & 2 \\
Beam & 20 & 2 & \{1,2,3\}\\
Paperclip & 576 & $<16$ & \{1,2,4,8,16,32,64\}\\
MNIST & $28^2$ & $\approx 10$ & \{1,2,4,8,16,32,64\} \\
Office & $3\times 128^2$ & 1 & \{1,2,3\} \\
\hline
\end{tabular}
\caption{Characteristics of all data sets and all latent space dimensions used in the experiments.}
\label{tab:characteristics}
\end{table}

The four autoencoder architectures of Table~\ref{tab:architectures} are evaluated on these test cases with different latent dimensions $d$. For all synthetic manifold and computational mechanics test cases, the nonlinear component of the architecture consists of a three-layer fully-connected neural network with ELU activations and 128 hidden units per layer. For image processing test cases, nonlinear encoders consist of 3 convolutional layers followed by 3 fully-connected layers, while decoders consist of 3 fully-connected layers followed by 3 transposed convolutional layers, with ReLU activations. No hyperparameter optimization was performed, as the goal of this work is not to find the optimal architecture for each test case, but to compare architectures against one another under consistent conditions.

To compare the four autoencoder architectures (LAE, Lenc-AE, Ldec-AE, AE), we trained models for each data set using the latent dimensions $d$ specified in Table~\ref{tab:characteristics}. In cases where $d\leq3$, the latent spaces were visualized and studied. All models were trained by minimizing mean squared error (MSE) reconstruction loss. For the Office data set, along with standard training results, an additional one, using a UMAP loss to structure the latent space, is provided. We intentionally avoided sophisticated training techniques, as the focus of this study is not on optimizing training procedures. Instead, we selected a sufficient number of epochs (typically 1,000) and a standard learning rate (either 0.01 or 0.001) to ensure convergence of the Adam optimizer. This approach ensures that each model achieves stable reconstruction performance, enabling a fair comparison of their intrinsic capabilities.

%% file: sections/04-results.tex
\section{Test-cases and Results}\label{sec:results}

\subsection{Synthetic Data Sets}

\input{sections/results/01-helix}

\input{sections/results/02-knot}

\input{sections/results/03-s-curve}

\input{sections/results/04-swiss-roll}

\subsection{Computational Mechanics Data Sets}

\input{sections/results/05-beam}

\input{sections/results/06-paperclip}

\subsection{Image Data Sets}
\input{sections/results/07-mnist}
\input{sections/results/08-office-video}

%% file: sections/results/01-helix.tex
\subsubsection{Helix}
The Helix test case consists of data points in $\mathbb{R}^3$ lying on a parametrized curve defined as:  

\begin{equation*}
\gamma(x) = \big(\sin(5x), \cos(5x), x\big), \quad x \in [-1, 1]
\end{equation*}

Since the data are distributed along an open curve, the intrinsic dimension of the test case is, by construction, equal to $1$: the manifold is homeomorphic to a closed interval of $\mathbb{R}$ and can be learnt by any nonlinear encoder with $d \geq 1$.

Moreover, this data set can theoretically be learned perfectly by a linear-encoder autoencoder (Lenc-AE) with $d=1$. Indeed, there exist lines on which the projection of the curve points is injective, for instance the helix’s central axis spanned by $(0,0,1)$. A nonlinear decoder can then learn the reciprocal of this projection.  

Concretely, one can define:  $\enc(X) =  (0,0,1)^\top X = x$ and $\dec(x) = \gamma(x)$. 

Therefore, with a sufficiently expressive nonlinear decoder, AE and Lenc-AE with any $d \geq 1$ can theoretically learn this data set perfectly. However, there is no guarantee that this can always be achieved in practice.

\begin{figure}[h]
\centering
\def\mywidth{24mm}
\begin{tabular}{m{2cm}m{2cm}m{2cm}m{2cm}m{2cm}m{2.cm}}
&AE&Lenc-AE&Ldec-AE&LAE&Objective\\\hline
$d=1$ &
\includegraphics[trim={16mm 16mm 16mm 16mm},clip,width=\mywidth]{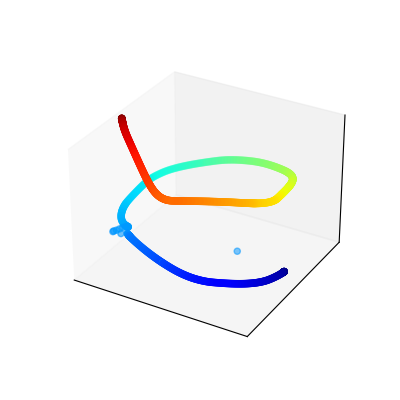}
&
\includegraphics[trim={16mm 16mm 16mm 16mm},clip,width=\mywidth]{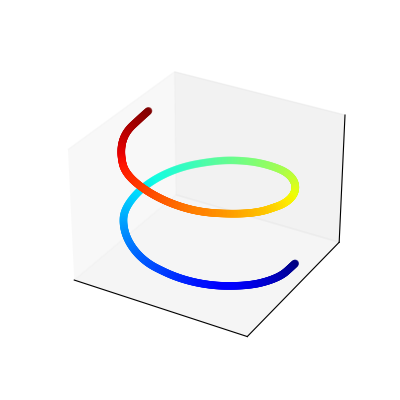} 
& \includegraphics[trim={16mm 16mm 16mm 16mm},clip,width=\mywidth]{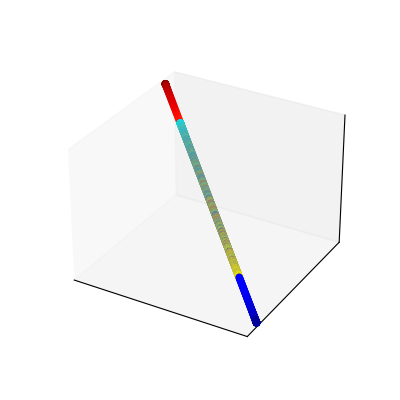} 
&\includegraphics[trim={16mm 16mm 16mm 16mm},clip,width=\mywidth]{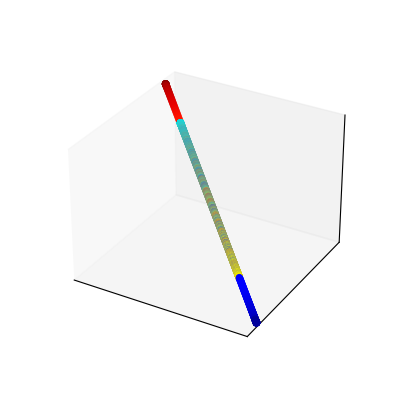} 
& \includegraphics[trim={16mm 16mm 16mm 16mm},clip,width=\mywidth]{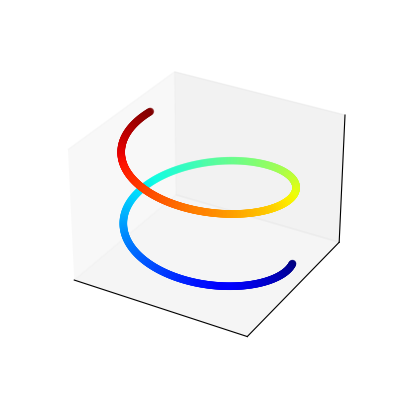} \\\hline
$d=2$ &
\includegraphics[trim={16mm 16mm 16mm 16mm},clip,width=\mywidth]{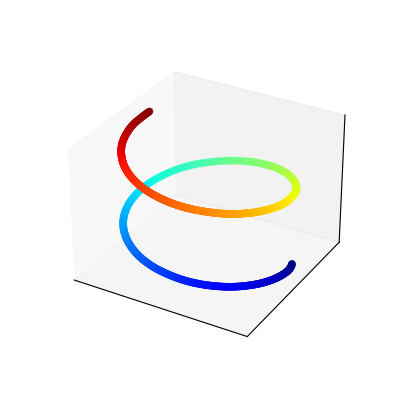}
&
\includegraphics[trim={16mm 16mm 16mm 16mm},clip,width=\mywidth]{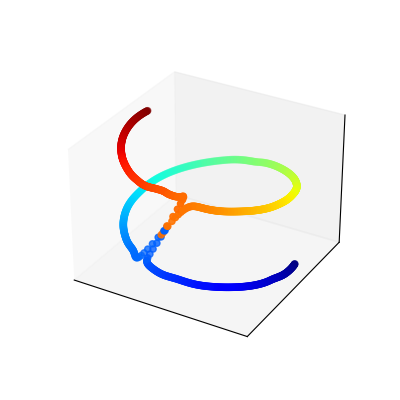} 
& \includegraphics[trim={16mm 16mm 16mm 16mm},clip,width=\mywidth]{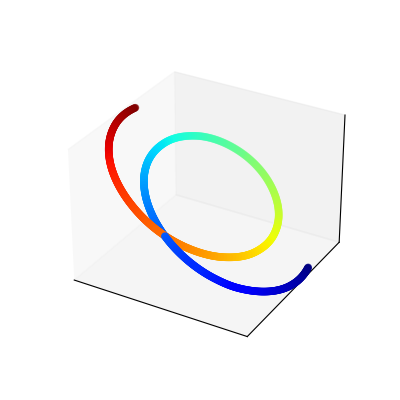} 
&\includegraphics[trim={16mm 16mm 16mm 16mm},clip,width=\mywidth]{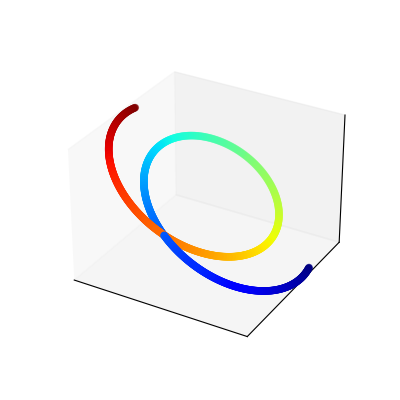} 
& \includegraphics[trim={16mm 16mm 16mm 16mm},clip,width=\mywidth]{helix/ref.png} \\
latent space &
\includegraphics[width=\mywidth]{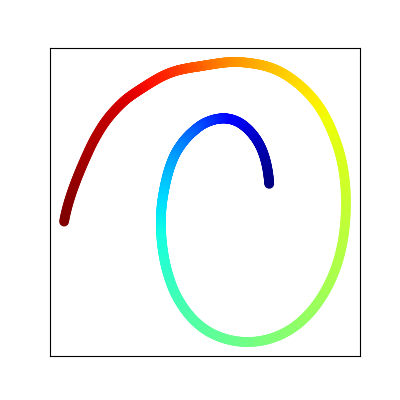}
&
\includegraphics[width=\mywidth]{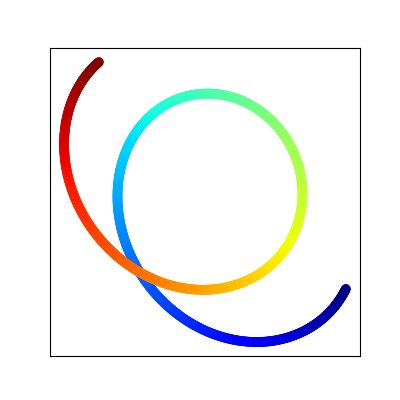} 
& \includegraphics[width=\mywidth]{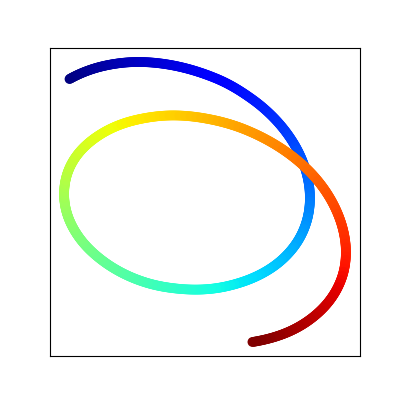} 
&\includegraphics[width=\mywidth]{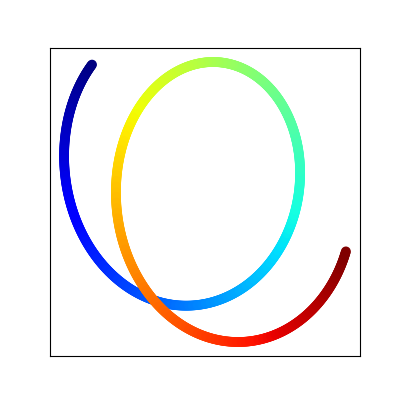} 
& \\
\hline
\end{tabular}
\caption{Each image displays the full helix data set, where each point is reconstructed using the four autoencoders. The corresponding latent representations are also provided for $d=2$.}\label{fig:plot_helix}
\end{figure}

We trained the four autoencoder architectures with latent spaces of dimension $1$ and $2$.  
Figure~\ref{fig:plot_helix} shows the reconstructed manifold in $\mathbb{R}^3$ (i.e., the curve $X \mapsto \dec(\enc(X))$) for the four architectures.  
For the experiments in dimension $2$, we also display the latent space representation, i.e., the curve $x \mapsto \enc(x)$. The mean squared errors (MSE) are provided in the general results table (Table~\ref{tab:reconstruction_errors}).

In both dimensions $1$ and $2$, the Ldec-AE and LAE are too limited in expressivity to reconstruct the manifold.  
The reconstructed manifold produced by these architectures can only lie within a line or a plane, which prevents an accurate recovery of the helix structure.  

In contrast, the Lenc-AE learns the manifold perfectly in dimension $1$.  It even outperforms the standard AE. With a more adapted training method inspired by the literature~\citep{DBLP:journals/corr/abs-2006-09289, Braunsmann_2024, umapae}, the nonlinear AE should also be able to recover the manifold perfectly for $d=1$, but none of this was needed to make the Lenc-AE learn it perfectly.

In dimension $2$, the Lenc-AE does not reconstruct the manifold perfectly, unlike the AE.  
This limitation comes from the fact that the projection into the latent space is not injective at certain points, which results in a discontinuity in the reconstruction. 

%% file: sections/results/02-knot.tex
\subsubsection{Knot}

The Knot test case consists of data points in $\mathbb{R}^3$ lying on a parametrized curve defined as:  
\begin{equation*}
\gamma(t) = \big(\cos(t) + 2\cos(2t), \; \sin(t) - 2\sin(2t), \; 2\sin(3t)\big), \quad t \in [0, 2\pi]
\end{equation*}

Since $\gamma(0) = \gamma(2\pi)$, this continuous curve is closed, forming a knot—a one-dimensional manifold embedded in $\mathbb{R}^3$ that is not homeomorphic to $\mathbb{R}$. Hence, this manifold cannot be perfectly learned with $d=1$ by a linear-encoder autoencoder.

According to Whitney's embedding theorem, a one-dimensional manifold can be embedded in $\mathbb{R}^2$ without self-intersections, but this requires a nonlinear projection. Therefore, only a nonlinear AE with $d \geq 2$ may be able to learn the manifold perfectly.

\begin{figure}[h]
\centering
\def\mywidth{24mm}
\begin{tabular}{m{2cm}m{2cm}m{2cm}m{2cm}m{2cm}m{2.cm}}
&AE&Lenc-AE&Ldec-AE&LAE&Objective\\\hline
$d=1$ &
\includegraphics[trim={16mm 16mm 16mm 16mm},clip,width=\mywidth]{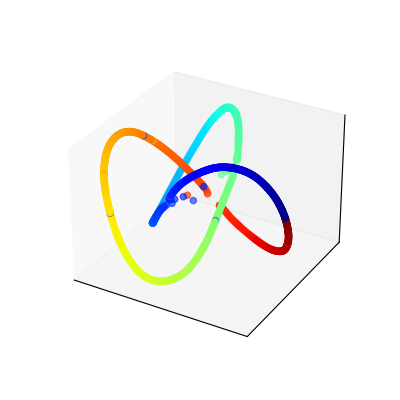}
&
\includegraphics[trim={16mm 16mm 16mm 16mm},clip,width=\mywidth]{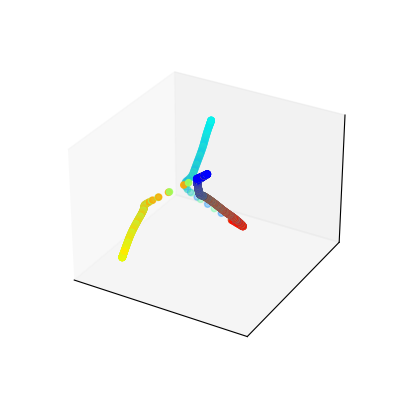} 
& \includegraphics[trim={16mm 16mm 16mm 16mm},clip,width=\mywidth]{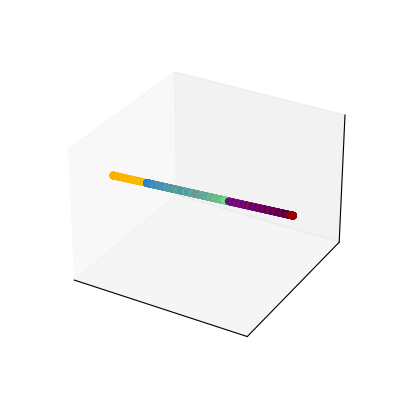} 
&\includegraphics[trim={16mm 16mm 16mm 16mm},clip,width=\mywidth]{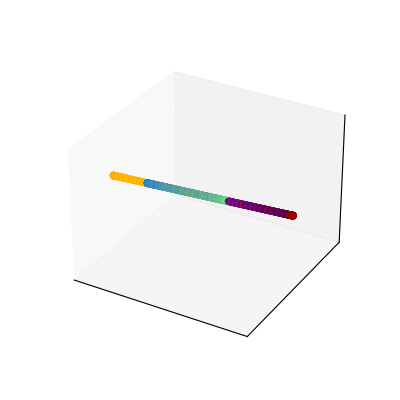} 
& \includegraphics[trim={16mm 16mm 16mm 16mm},clip,width=\mywidth]{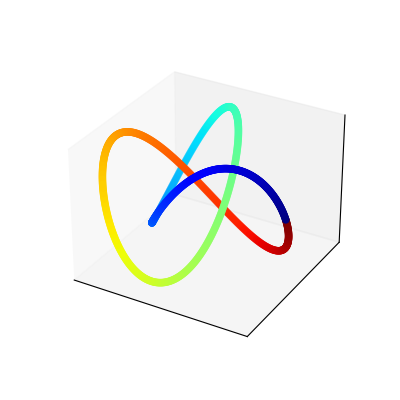} \\
\hline
$d=2$ &
\includegraphics[trim={7mm 7mm 7mm 7mm},clip,width=\mywidth]{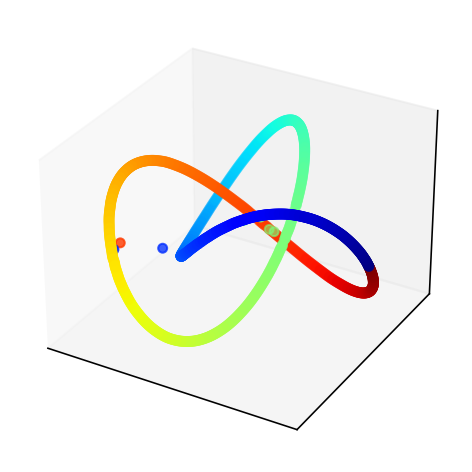}
&
\includegraphics[trim={7mm 7mm 7mm 7mm},clip,width=\mywidth]{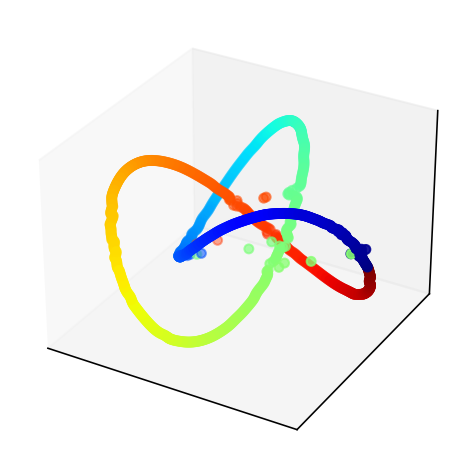} 
& \includegraphics[trim={7mm 7mm 7mm 7mm},clip,width=\mywidth]{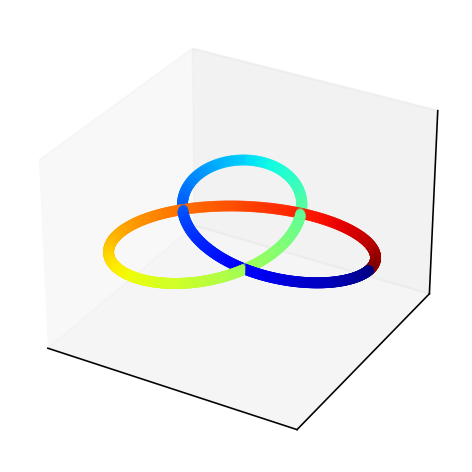} 
&\includegraphics[trim={7mm 7mm 7mm 7mm},clip,width=\mywidth]{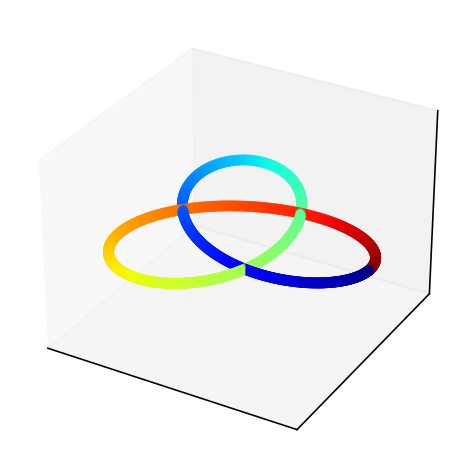} 
& \includegraphics[trim={16mm 16mm 16mm 16mm},clip,width=\mywidth]{knot/ref.png} \\
latent space &
\includegraphics[width=\mywidth]{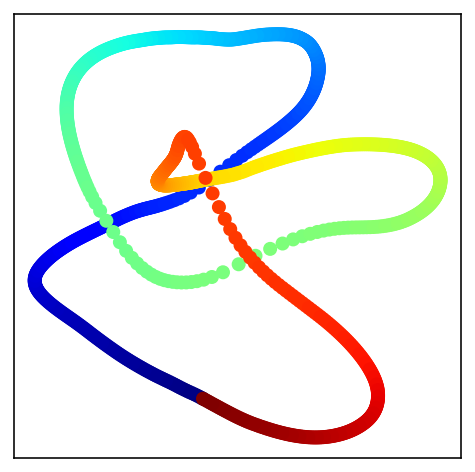}
&
\includegraphics[width=\mywidth]{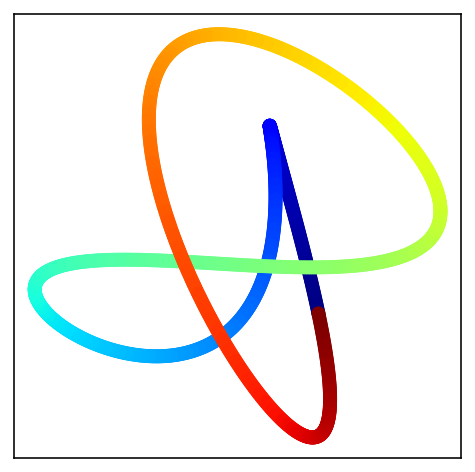} 
& \includegraphics[width=\mywidth]{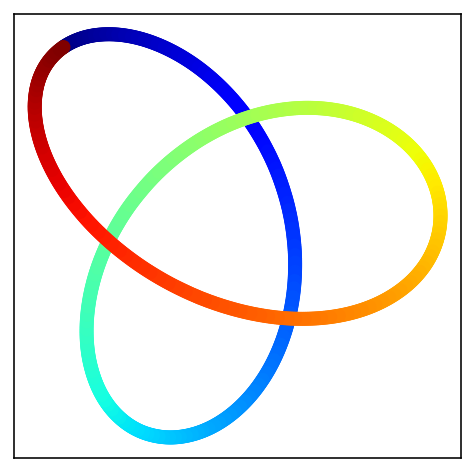} 
&\includegraphics[width=\mywidth]{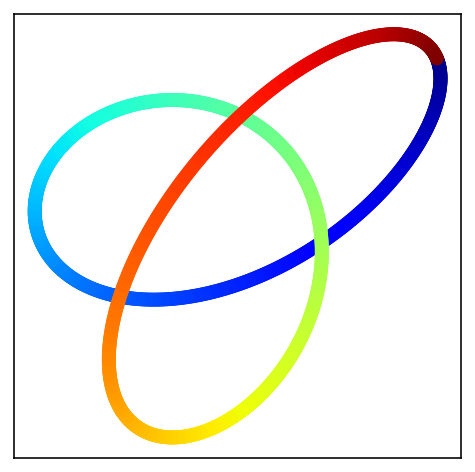} 
& \\
\hline
\end{tabular}
\caption{Each image displays the full knot data set, where each point is reconstructed using the four autoencoders. The corresponding latent representations are also provided for $d=2$.}\label{fig:plot_knot}
\end{figure}

In practice however, minimizing reconstruction MSE alone is insufficient to guarantee a self-intersection-free embedding: the loss provides no injectivity constraint, and distinct parts of the knot can collapse to the same latent coordinates. Achieving a proper embedding would likely require a dedicated training strategy, such as a topology-aware loss~\citep{moor2021topologicalautoencoders}. Results are depicted in Figure~\ref{fig:plot_knot}.

As expected, the LAE and Ldec-AE fail to approximate the knot: their linear decoders constrain the output to lines or planes, geometries fundamentally incapable of representing the knot's intricate topology.
The AE achieves partial reconstructions for both $d=1$ and $d=2$, recovering the global shape of the knot but with discontinuities. These arise from the non-injectivity of the encoder in certain regions, where distinct parts of the knot are mapped to the same latent coordinates. 

The Lenc-AE behaves similarly for $d=2$, recovering the global structure with comparable discontinuities. For $d=1$, its linear encoder further restricts representational capacity, preventing any meaningful approximation.

While partial reconstructions may suffice for visualization tasks, the discontinuities pose a critical limitation for applications requiring smoothness, such as solving ODEs on the learned manifold.

%% file: sections/results/03-s-curve.tex
\subsubsection{S-curve}

The S-curve data set consists of points in $\mathbb{R}^3$ lying on a two-dimensional manifold parametrized as:

\begin{equation*}
\gamma(x, y) = \left(\sin(x), 2y, \mathrm{sign}(x)(\cos(x) - 1) \right), \quad x \in [-3\pi /2, 3\pi /2], \quad y \in [0,1].
\end{equation*}
This manifold is homeomorphic to a planar rectangle, meaning it can be continuously flattened into $\mathbb{R}^2$
 without self-intersections. However, its curved geometry implies that there is no linear mapping from 
$\mathbb{R}^3$ to $\mathbb{R}^2$ that is injective when restricted to the S-curve: any linear projection will inevitably fold or overlap parts of the curve, failing to preserve its intrinsic structure.

\begin{figure}[h]
\centering
\def\mywidth{24mm}
\begin{tabular}{m{2cm}m{2cm}m{2cm}m{2cm}m{2cm}m{2.cm}}
&AE&Lenc-AE&Ldec-AE&LAE&Objective\\\hline
$d=2$ &
\includegraphics[trim={16mm 16mm 16mm 16mm},clip,width=\mywidth]{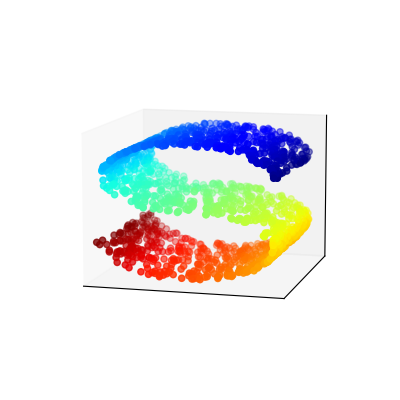}
&
\includegraphics[trim={16mm 16mm 16mm 16mm},clip,width=\mywidth]{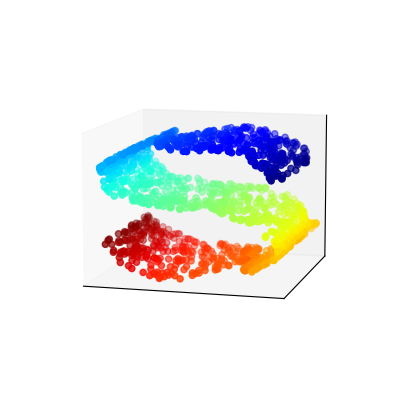} 
& \includegraphics[trim={16mm 16mm 16mm 16mm},clip,width=\mywidth]{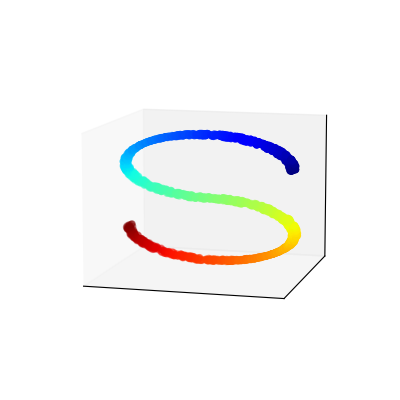} 
&\includegraphics[trim={16mm 16mm 16mm 16mm},clip,width=\mywidth]{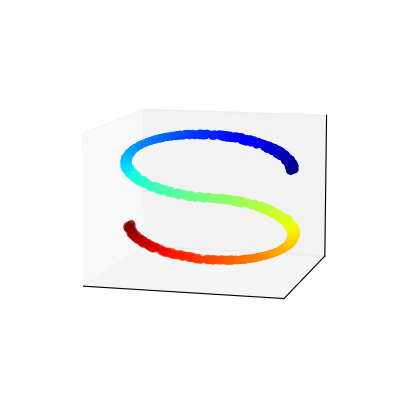} 
& \includegraphics[trim={16mm 16mm 16mm 16mm},clip,width=\mywidth]{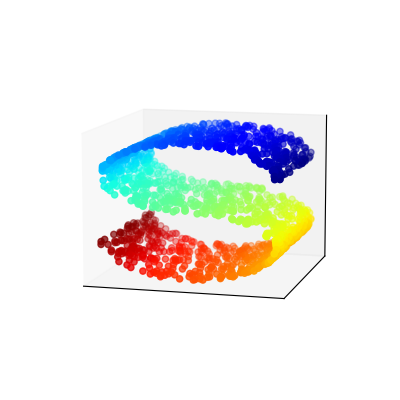} \\
latent space &
\includegraphics[width=\mywidth]{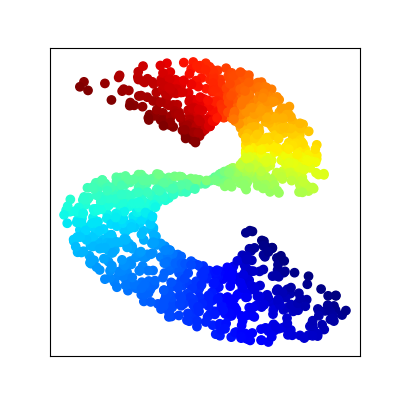}
&
\includegraphics[width=\mywidth]{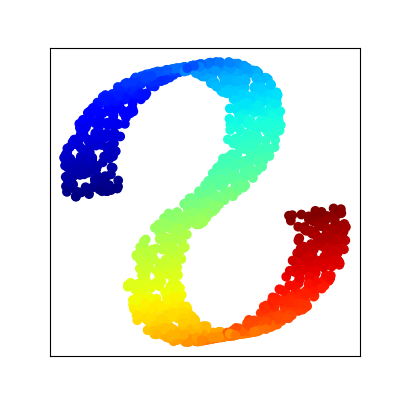} 
& \includegraphics[width=\mywidth]{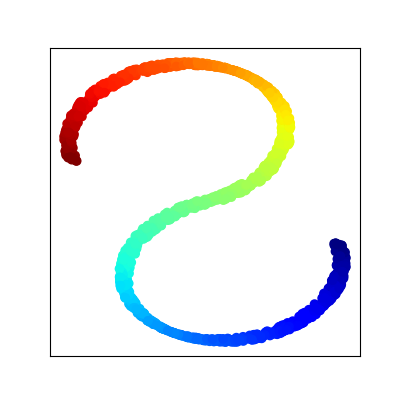} 
&\includegraphics[width=\mywidth]{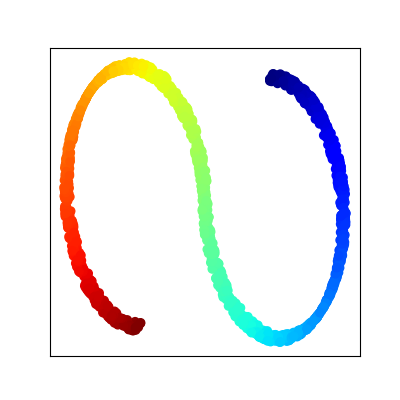} 
& \\
\hline
\end{tabular}
\caption{Each image displays the full S-curve data set, where each point is reconstructed using the four autoencoders. The corresponding latent representations are also provided.}\label{fig:Scurve}
\end{figure}

Results are depicted in Figure~\ref{fig:Scurve}. As expected, the Lenc-AE is therefore less accurate than AE in reconstructing the S-curve, since no linear projection can avoid overlapping regions entirely. 
Both architectures outperform the Ldec-AE and LAE, which fail to capture the curved geometry due to their linear decoders, confining their outputs to a planar subspace when $d=2$ or to a straight line when $d=1$.

%% file: sections/results/04-swiss-roll.tex
\subsubsection{Swiss Roll}

The Swiss roll data set consists of points in $\mathbb{R}^3$ lying on a two-dimensional manifold parametrized as:

\begin{equation*}
\gamma(x, y) = \big(x \cos(x), x \sin(x), y\big), \quad x \in [0, 3\pi], \quad y \in [-1, 1].
\end{equation*}

Similarly, the Swiss roll is homeomorphic to a rectangle, but its rolled geometry prevents any linear projection from preserving its structure without overlaps.

\begin{figure}[h]
\centering
\def\mywidth{24mm}
\begin{tabular}{m{2cm}m{2cm}m{2cm}m{2cm}m{2cm}m{2.cm}}
&AE&Lenc-AE&Ldec-AE&LAE&Objective\\\hline
$d=2$ &
\includegraphics[trim={16mm 16mm 16mm 16mm},clip,width=\mywidth]{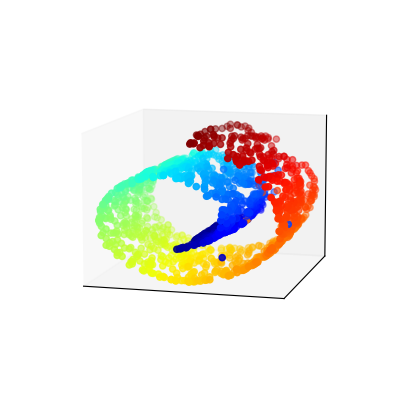}
&
\includegraphics[trim={16mm 16mm 16mm 16mm},clip,width=\mywidth]{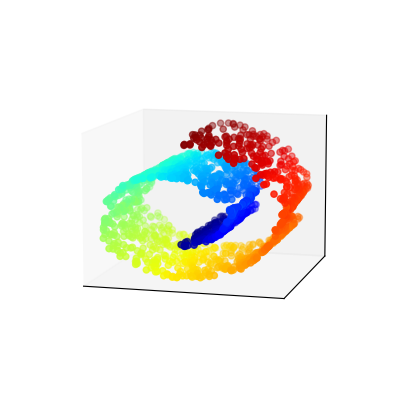} 
& \includegraphics[trim={16mm 16mm 16mm 16mm},clip,width=\mywidth]{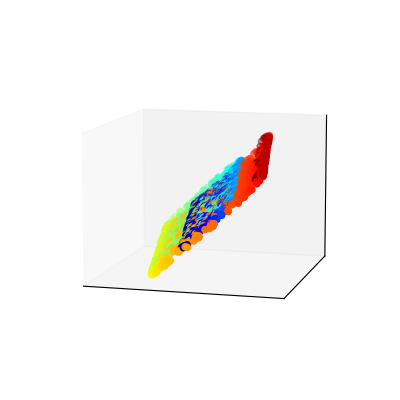} 
&\includegraphics[trim={16mm 16mm 16mm 16mm},clip,width=\mywidth]{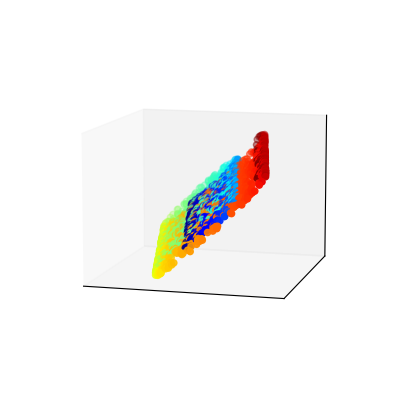} 
& \includegraphics[trim={16mm 16mm 16mm 16mm},clip,width=\mywidth]{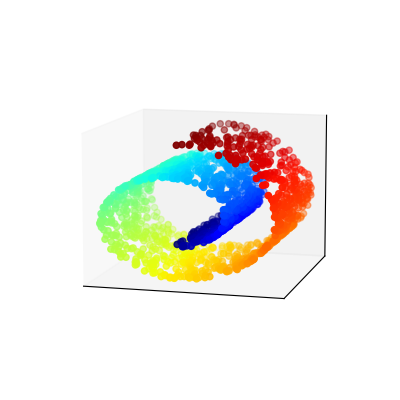} \\
latent space &
\includegraphics[width=\mywidth]{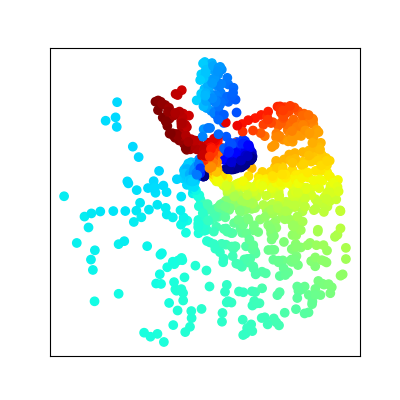}
&
\includegraphics[width=\mywidth]{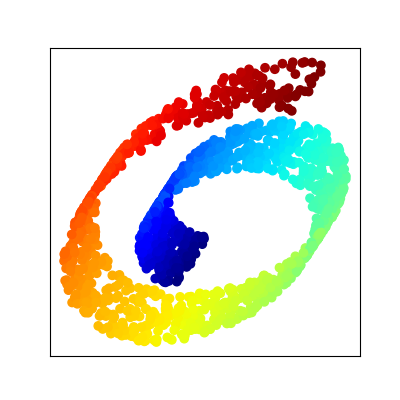} 
& \includegraphics[width=\mywidth]{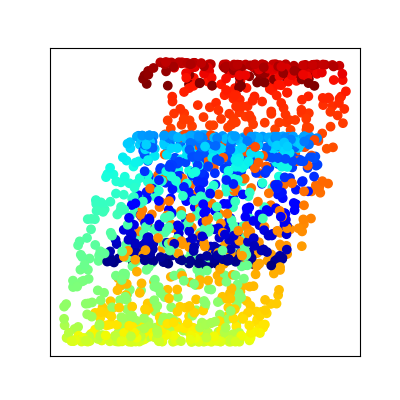} 
&\includegraphics[width=\mywidth]{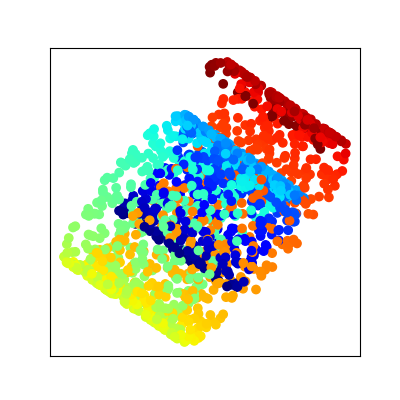} 
& \\
\hline
\end{tabular}
\caption{Each image displays the full Swiss roll data set, where each point is reconstructed using the four autoencoders. The corresponding latent representations are also provided.}\label{fig:swissroll}
\end{figure}

Results are depicted in Figure~\ref{fig:swissroll} for $d=2$. As with the S-curve, the Ldec-AE and LAE remain unable to capture the nonlinear geometry, their outputs confined to two-dimensional linear subspaces that cannot represent the rolled structure. 

The AE accurately reconstructs the Swiss roll, but without specific regularization techniques its latent space bears little resemblance to the original structure, as the nonlinear encoder learns an arbitrary mapping optimized for reconstruction.

The Lenc-AE, constrained by its linear encoder, produces a more organized latent space that reflects the global geometry of the Swiss roll. This comes at the cost of reconstruction accuracy: in regions where the linear projection fails to be injective, distinct parts of the Swiss roll are mapped to the same latent coordinates, resulting in discontinuities and distortions in the decoded manifold.

%% file: sections/results/05-beam.tex
\subsubsection{Beam}
\begin{figure}[h]
\centering
\def\mywidth{24mm}
\begin{tabular}{m{2cm}m{2cm}m{2cm}m{2cm}m{2cm}m{2.cm}}
&AE&Lenc-AE&Ldec-AE&LAE&Objective\\\hline
$d=1$ &
\includegraphics[trim={14mm 14mm 14mm 14mm},clip,width=\mywidth]{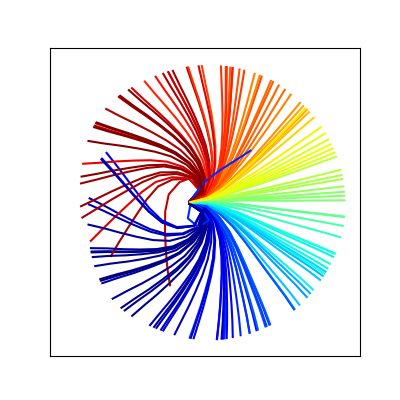}
&
\includegraphics[trim={14mm 14mm 14mm 14mm},clip,width=\mywidth]{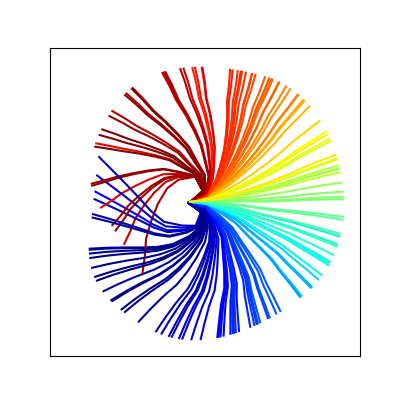} 
& \includegraphics[trim={14mm 14mm 14mm 14mm},clip,width=\mywidth]{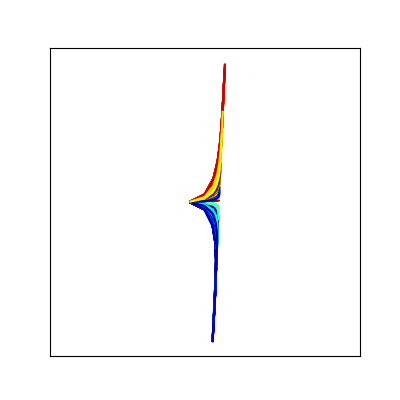} 
&\includegraphics[trim={14mm 14mm 14mm 14mm},clip,width=\mywidth]{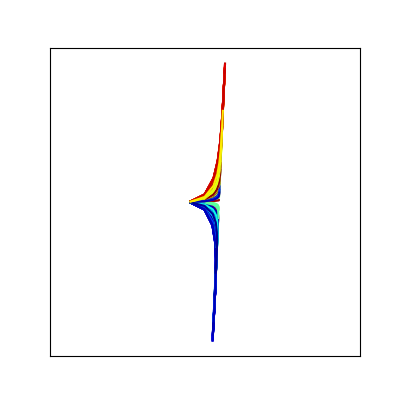} 
& \includegraphics[trim={14mm 14mm 14mm 14mm},clip,width=\mywidth]{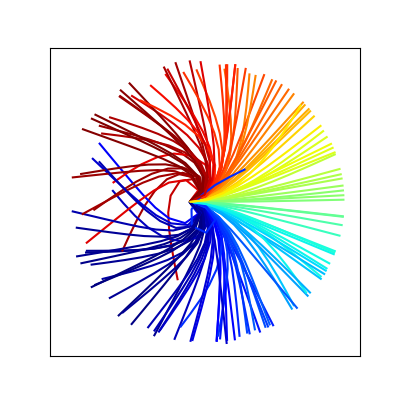} \\
\hline
$d=2$ &
\includegraphics[trim={14mm 14mm 14mm 14mm},clip,width=\mywidth]{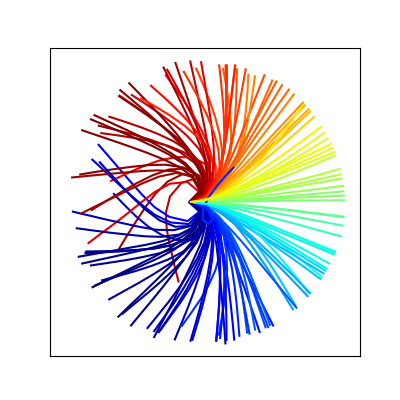}
&
\includegraphics[trim={14mm 14mm 14mm 14mm},clip,width=\mywidth]{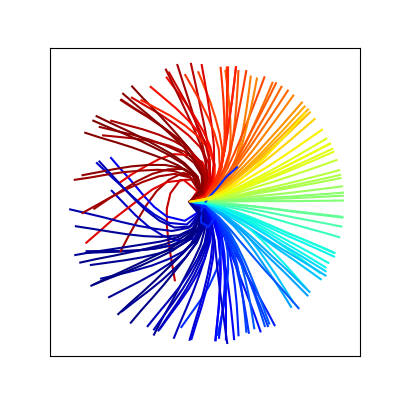} 
& \includegraphics[trim={14mm 14mm 14mm 14mm},clip,width=\mywidth]{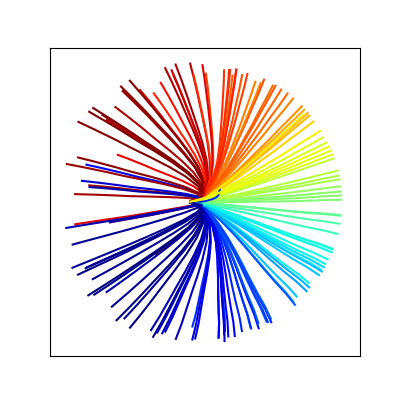} 
&\includegraphics[trim={14mm 14mm 14mm 14mm},clip,width=\mywidth]{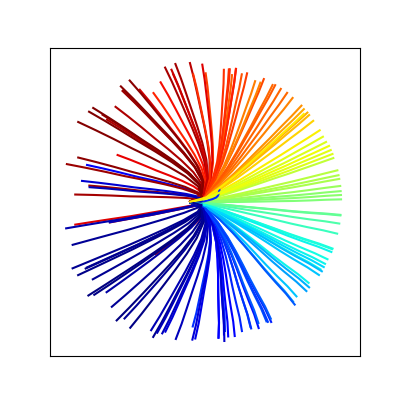} 
& \includegraphics[trim={14mm 14mm 14mm 14mm},clip,width=\mywidth]{beam/ref.png} \\
latent space &
\includegraphics[width=\mywidth]{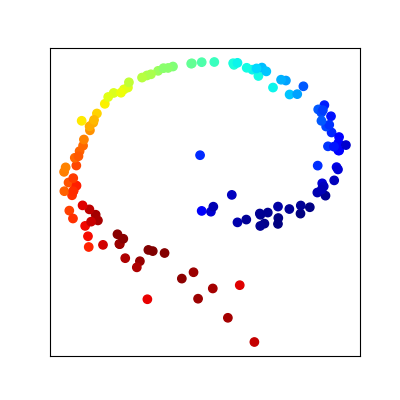}
&
\includegraphics[width=\mywidth]{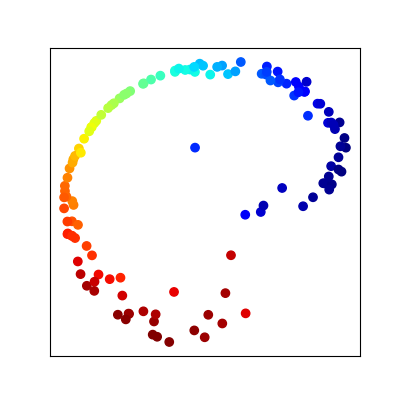} 
& \includegraphics[width=\mywidth]{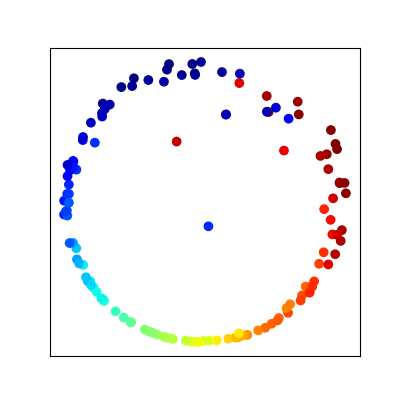} 
&\includegraphics[width=\mywidth]{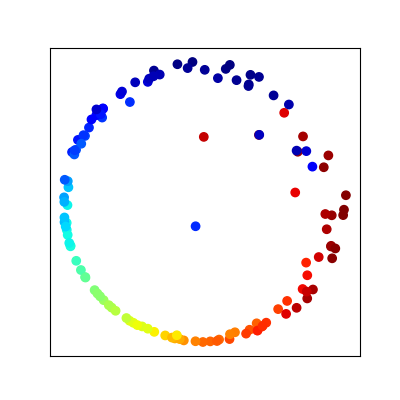} 
& \\
\hline
$d=3$ &
\includegraphics[trim={14mm 14mm 14mm 14mm},clip,width=\mywidth]{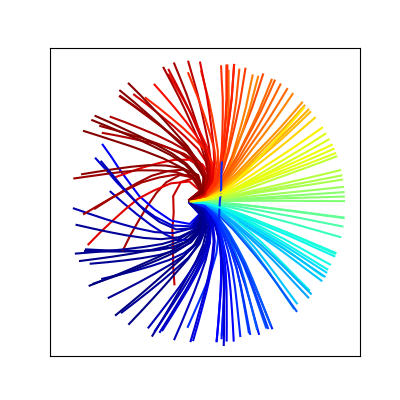}
&
\includegraphics[trim={14mm 14mm 14mm 14mm},clip,width=\mywidth]{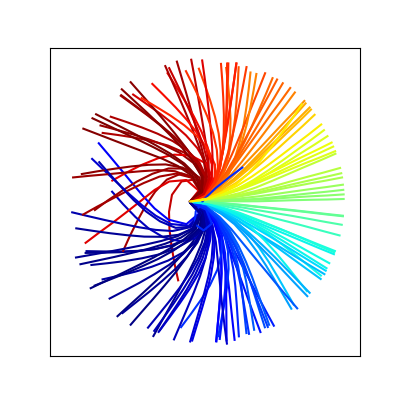} 
& \includegraphics[trim={14mm 14mm 14mm 14mm},clip,width=\mywidth]{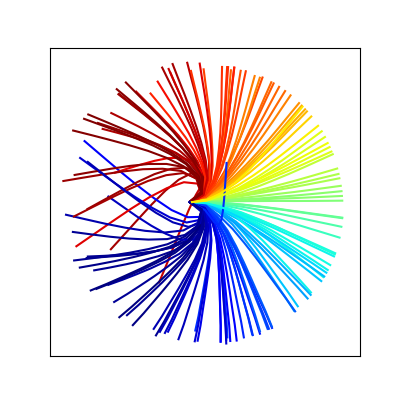} 
&\includegraphics[trim={14mm 14mm 14mm 14mm},clip,width=\mywidth]{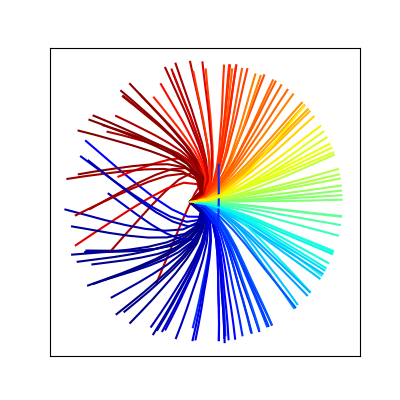} 
& \includegraphics[trim={14mm 14mm 14mm 14mm},clip,width=\mywidth]{beam/ref.png} \\
latent space &
\includegraphics[trim={16mm 16mm 16mm 16mm},clip,width=\mywidth]{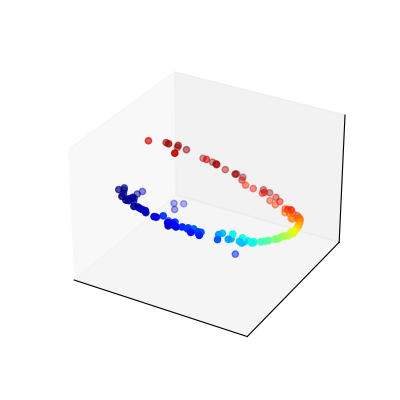}
&
\includegraphics[trim={16mm 16mm 16mm 16mm},clip,width=\mywidth]{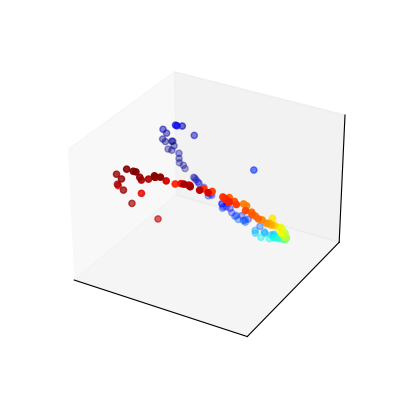} 
& \includegraphics[trim={16mm 16mm 16mm 16mm},clip,width=\mywidth]{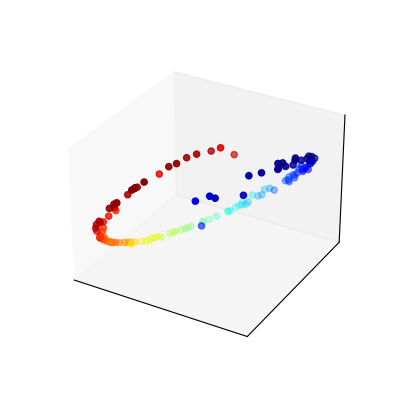} 
&\includegraphics[trim={16mm 16mm 16mm 16mm},clip,width=\mywidth]{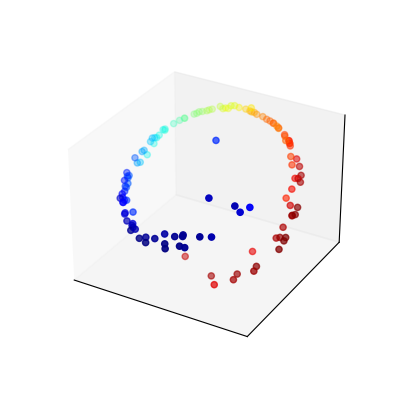} 
& \\
\hline
\end{tabular}
\caption{Each image depicts a collection of deformed beams reconstructed by each architecture. For $d\in\{2,3\}$, the latent space visualization is also provided. The color of each dot, representing a data point, matches the color of the corresponding deformed beam.}\label{fig:plot_beam}
\end{figure}

This data set consists of displacement fields of a nonlinear beam subjected to a 2D load at its tip. The beam is discretized using the finite element method with 10 nodes, yielding displacement fields of 20 degrees of freedom. This benchmark and its detailed description are publicly available~\citep{LEBNN}.
The set of displacement fields obtained by varying a two-component load defines a manifold of dimension $p=2$. Whitney's theorem therefore guarantees that a latent space of dimension $d=4$ should be sufficient for a nonlinear AE to achieve a perfect reconstruction. In practice, our experiments show that $d=2$ already suffices for both the AE and Lenc-AE; results are presented in Figure~\ref{fig:plot_beam}.
For $d=1$, the AE and Lenc-AE learn a partial reconstruction corresponding to pure bending: the load component tangent to the beam induces displacements much smaller than those caused by the normal component, so the tangential contribution is effectively discarded. The Ldec-AE and LAE, whose decoders are too restrictive, fail to reconstruct meaningful displacement fields.

For $d=2$, the AE and Lenc-AE both achieve visually accurate reconstructions. Their latent spaces reveal that the topology of the problem has been correctly captured: points corresponding to upward and downward bending configurations (dark blue and dark red points) are clearly separated. This separation is absent in the latent spaces of the Ldec-AE and LAE, where these configurations are mixed together.

%% file: sections/results/06-paperclip.tex
\subsubsection{Paperclip}

The paperclip case is another example from computational mechanics. The original data set~\citep{HQJMYA_2025} contains 1,000 sequences of 300 steps of finite element computations of the deformation of a paperclip, undergoing a quasi-static action of two hands applying balanced wrench forces. The simulations are performed using a finite-strain elastoplastic constitutive law. The complex data generation procedure~\citep{lesueur:hal-05073095}, combined with the nonlinearity of elastoplasticity—which depends on the deformation history—makes it impossible to estimate the intrinsic dimension. For this reason, multiple values of the latent dimension $d$ were tested across the four architectures: 1, 2, 4, 8, 16, 32 and 64.

\begin{figure}[h]
\centering
\def\mywidth{28mm}
\begin{tabular}{m{2cm}m{2cm}m{2cm}m{2cm}m{2cm}m{2.cm}}
&AE&Lenc-AE&Ldec-AE&LAE&Objective\\\hline
$d=2$ &
\includegraphics[clip,width=\mywidth]{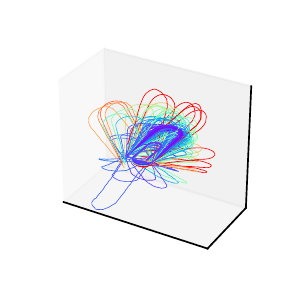}
&
\includegraphics[clip,width=\mywidth]{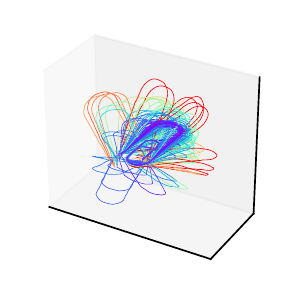} 
& \includegraphics[clip,width=\mywidth]{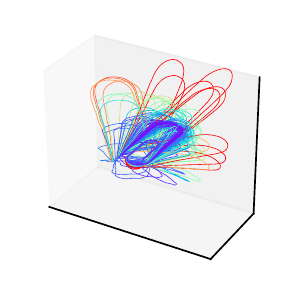}
&\includegraphics[clip,width=\mywidth]{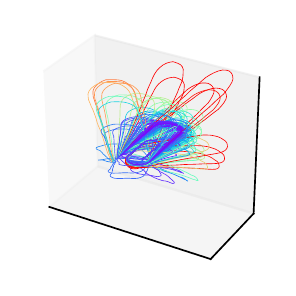}
& \includegraphics[clip,width=\mywidth]{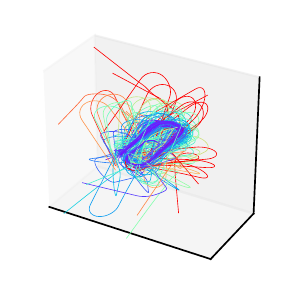} \\
\hline
$d=4$ &
\includegraphics[clip,width=\mywidth]{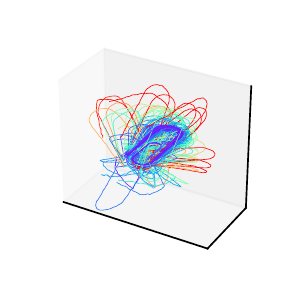}
&
\includegraphics[clip,width=\mywidth]{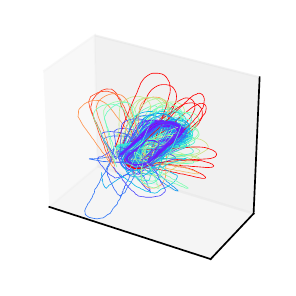} 
& \includegraphics[clip,width=\mywidth]{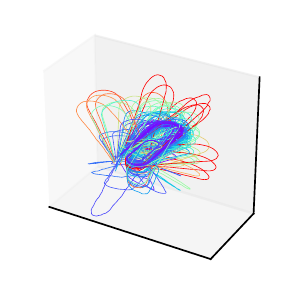}
&\includegraphics[clip,width=\mywidth]{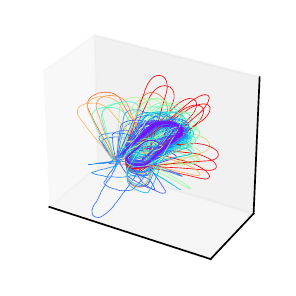}
& \includegraphics[clip,width=\mywidth]{paperclip/gt.pdf} \\
\hline
$d=8$ &
\includegraphics[clip,width=\mywidth]{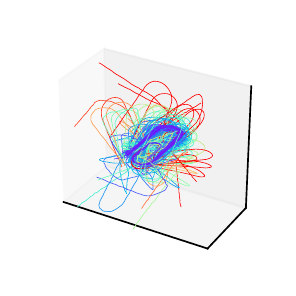}
&
\includegraphics[clip,width=\mywidth]{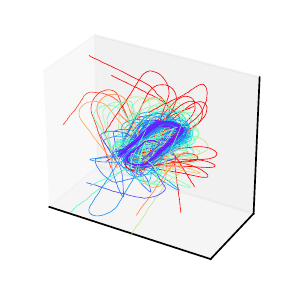} 
& \includegraphics[clip,width=\mywidth]{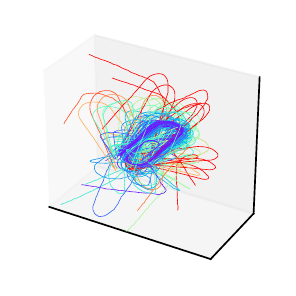}
&\includegraphics[clip,width=\mywidth]{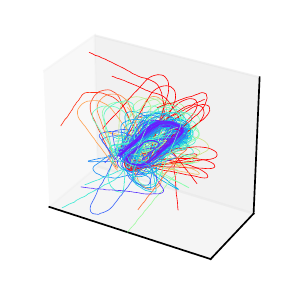}
& \includegraphics[clip,width=\mywidth]{paperclip/gt.pdf} \\
\hline
$d=16$ &
\includegraphics[clip,width=\mywidth]{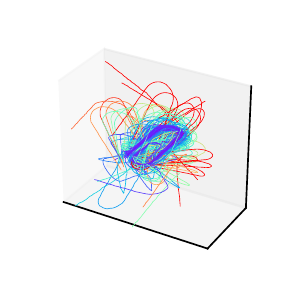}
&
\includegraphics[clip,width=\mywidth]{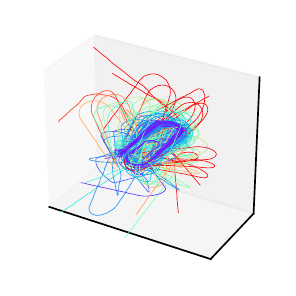} 
& \includegraphics[clip,width=\mywidth]{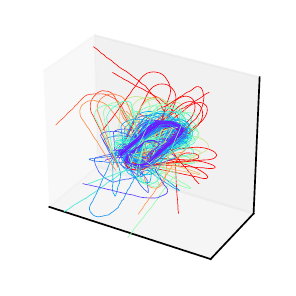}
&\includegraphics[clip,width=\mywidth]{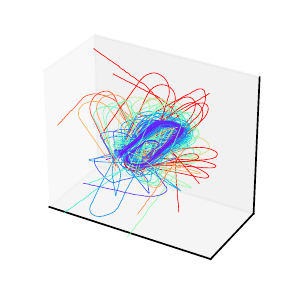}
& \includegraphics[clip,width=\mywidth]{paperclip/gt.pdf} \\
\hline
\end{tabular}
\caption{Each image depicts a collection of deformation steps of a paperclip, reconstructed by each architecture.}\label{tab:paperclip}
\end{figure}

Reconstruction metrics for $d\in\{1,2,4,8,16,32,64\}$ are reported in Table~\ref{tab:reconstruction_errors}; examples of reconstructions are shown in Figure~\ref{tab:paperclip}.

Reconstruction losses follow a consistent hierarchy across all values of $d$: the AE achieves the lowest error, followed by the Lenc-AE, while the Ldec-AE and LAE exhibit significantly higher and near-identical errors. Although the reconstruction error continues to decrease with $d$, results are almost indistinguishable from the target for all four architectures from $d=8$ onward.

%% file: sections/results/07-mnist.tex
\subsubsection{MNIST}

The MNIST data set, comprising $28 \times 28$ grayscale images of handwritten digits, serves as a benchmark for evaluating the performance of autoencoder architectures on high-dimensional, real-world data. While its intrinsic dimension $p$ is not known a priori, several studies have attempted to estimate it using diverse methodological approaches. For instance, the intrinsic dimension of image data sets was evaluated and led to the conclusion that for MNIST, $p$ lies between 7 and 13~\citep{pope2021intrinsicdimensionimagesimpact}. Other works have further refined this estimate by leveraging autoencoder-based compression techniques. An autoencoder was trained with a 128-dimensional latent space, followed by a linear transformation into another 128-dimensional space~\citep{jing2020implicitrankminimizingautoencoder}. By minimizing the rank of this linear transformation, the authors achieved low reconstruction errors with effective ranks below 10, suggesting that MNIST can be compressed into a latent space of dimension $p \leq 10$. Similarly, an additional compression step to $\mathbb{R}^9$ between two 128-dimensional layers was introduced~\citep{zeng2023autoencodersdiscoveringmanifolddimension}, leading to an estimation for the intrinsic dimension of MNIST to be $p = 9$. These findings align with the broader observation that MNIST’s manifold structure, while complex, can be effectively captured in a low-dimensional latent space.

To further explore the limits of dimensionality reduction for MNIST, an autoencoder with geodesic interpolation in the latent space was employed~\citep{krishnagopal2021encodedpriorslicedwasserstein}, achieving accurate reconstructions with a latent dimension as low as $d = 5$. This result suggests that the intrinsic dimension of MNIST may be even smaller than previously estimated, motivating further experiments with latent dimensions $d < 16$ to assess the trade-offs between reconstruction quality and latent space compactness.

\begin{figure}[h!]
\centering
\def\mywidth{24mm}
\begin{tabular}{ccccc}
AE&Lenc-AE&Ldec-AE&LAE&Objective\\
\hline

\includegraphics[trim={5mm 5mm 5mm 5mm},clip,width=\mywidth]{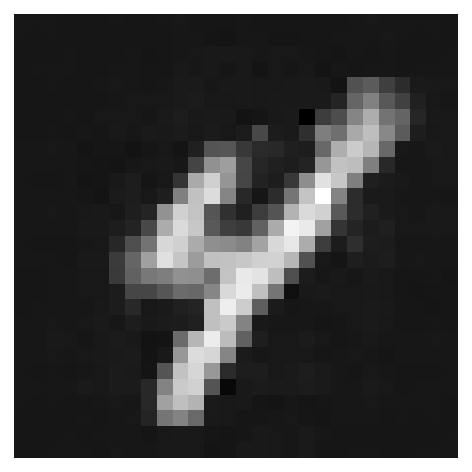}
& \includegraphics[trim={5mm 5mm 5mm 5mm},clip,width=\mywidth]{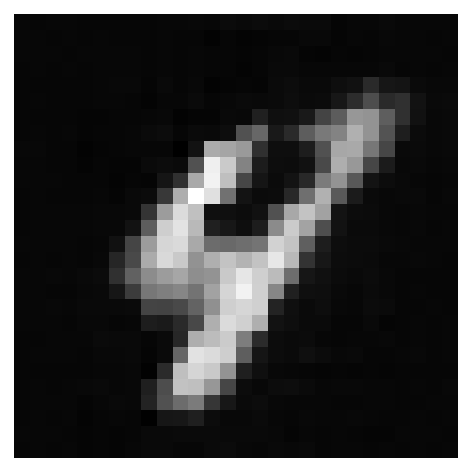}
&\includegraphics[trim={5mm 5mm 5mm 5mm},clip,width=\mywidth]{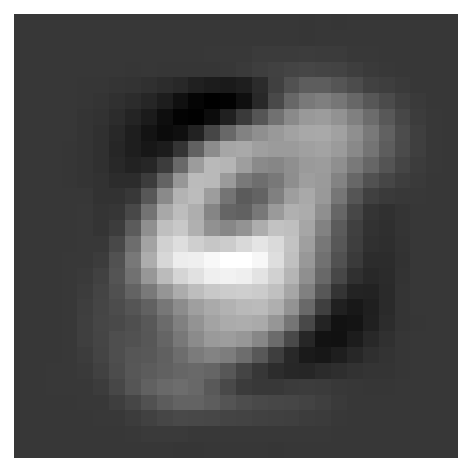}
& \includegraphics[trim={5mm 5mm 5mm 5mm},clip,width=\mywidth]{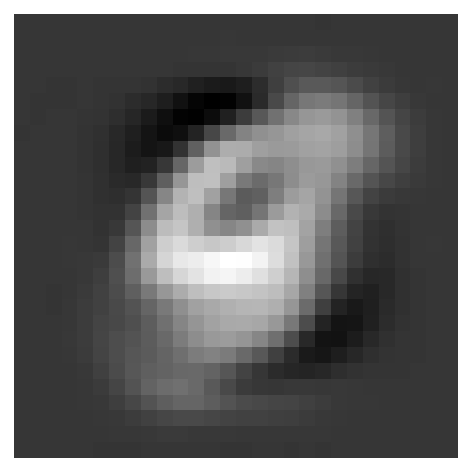}
&\includegraphics[trim={5mm 5mm 5mm 5mm},clip,width=\mywidth]{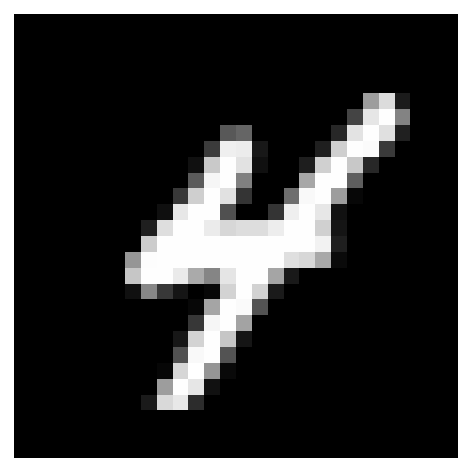}

 \\

\includegraphics[trim={5mm 5mm 5mm 5mm},clip,width=\mywidth]{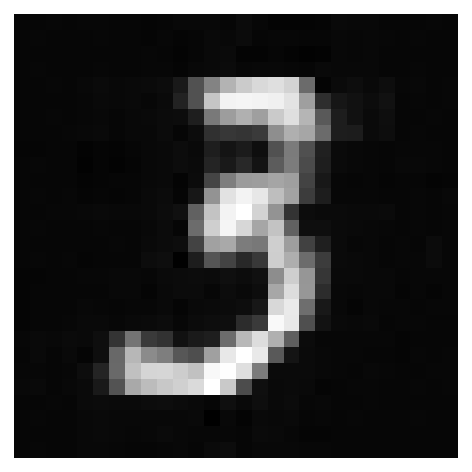}
& \includegraphics[trim={5mm 5mm 5mm 5mm},clip,width=\mywidth]{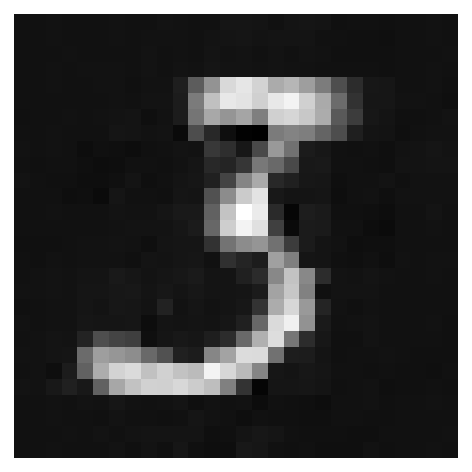}
&\includegraphics[trim={5mm 5mm 5mm 5mm},clip,width=\mywidth]{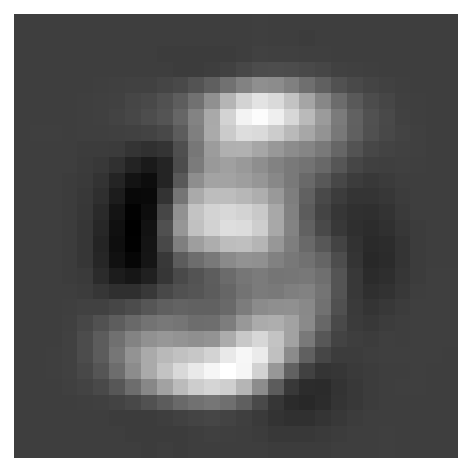}
& \includegraphics[trim={5mm 5mm 5mm 5mm},clip,width=\mywidth]{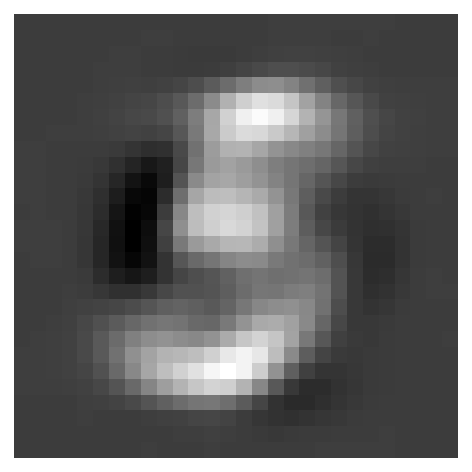}
&\includegraphics[trim={5mm 5mm 5mm 5mm},clip,width=\mywidth]{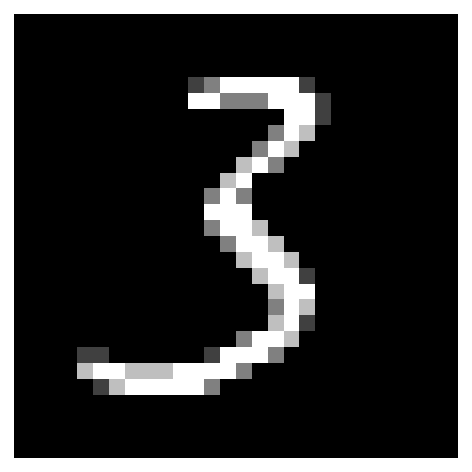}

 \\
 
 \includegraphics[trim={5mm 5mm 5mm 5mm},clip,width=\mywidth]{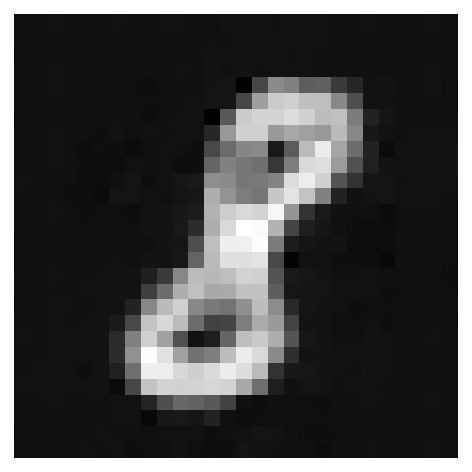}
& \includegraphics[trim={5mm 5mm 5mm 5mm},clip,width=\mywidth]{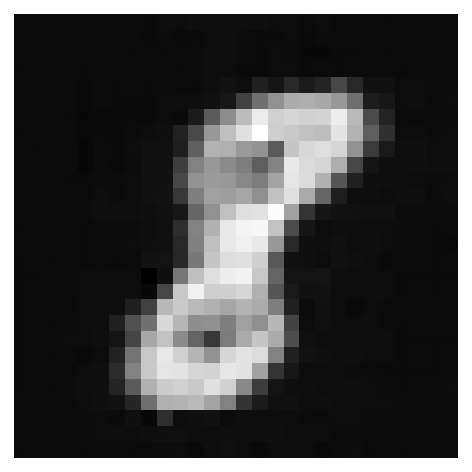}
&\includegraphics[trim={5mm 5mm 5mm 5mm},clip,width=\mywidth]{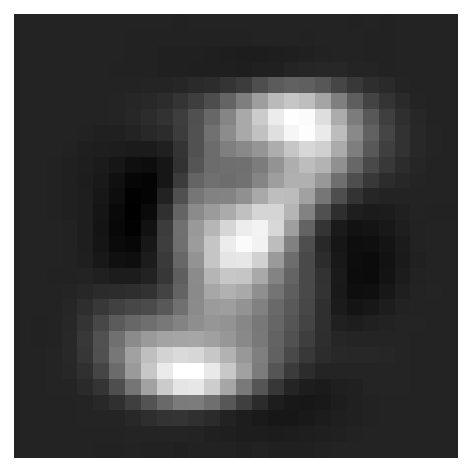}
& \includegraphics[trim={5mm 5mm 5mm 5mm},clip,width=\mywidth]{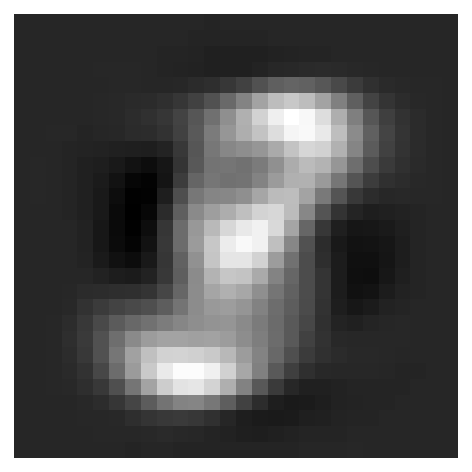}
&\includegraphics[trim={5mm 5mm 5mm 5mm},clip,width=\mywidth]{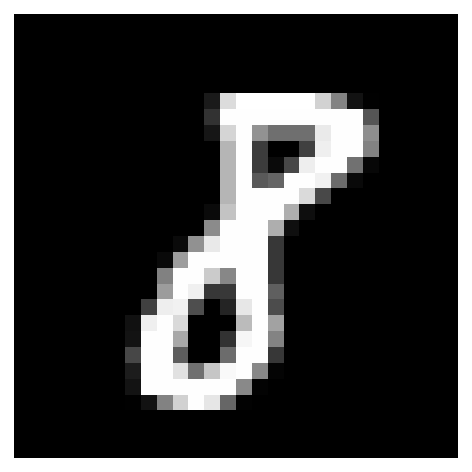}

 \\
 
 \includegraphics[trim={5mm 5mm 5mm 5mm},clip,width=\mywidth]{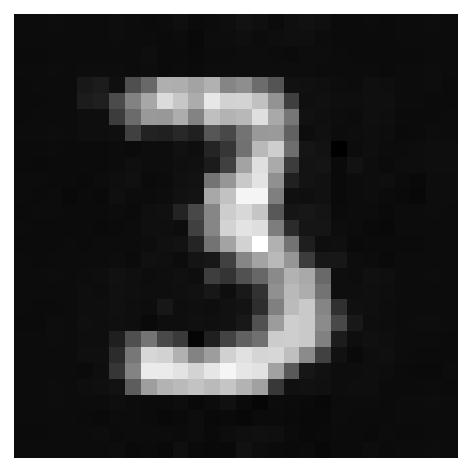}
& \includegraphics[trim={5mm 5mm 5mm 5mm},clip,width=\mywidth]{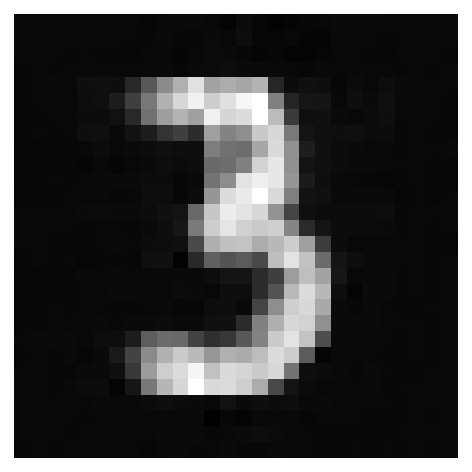}
&\includegraphics[trim={5mm 5mm 5mm 5mm},clip,width=\mywidth]{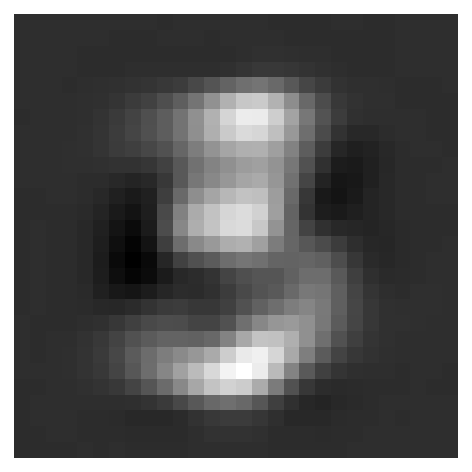}
& \includegraphics[trim={5mm 5mm 5mm 5mm},clip,width=\mywidth]{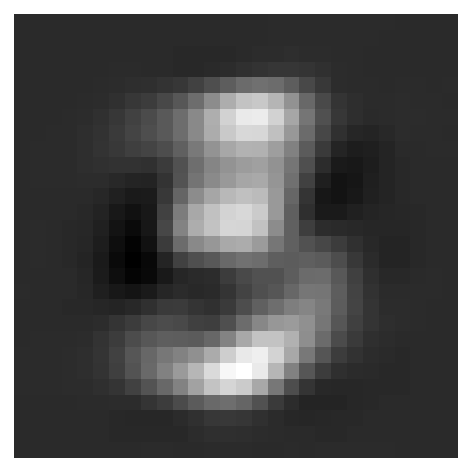}
&\includegraphics[trim={5mm 5mm 5mm 5mm},clip,width=\mywidth]{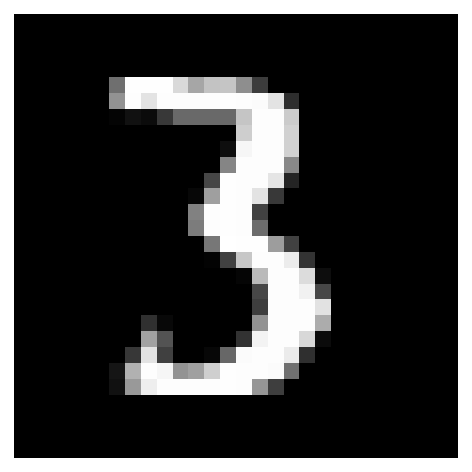}

 \\

  \includegraphics[trim={5mm 5mm 5mm 5mm},clip,width=\mywidth]{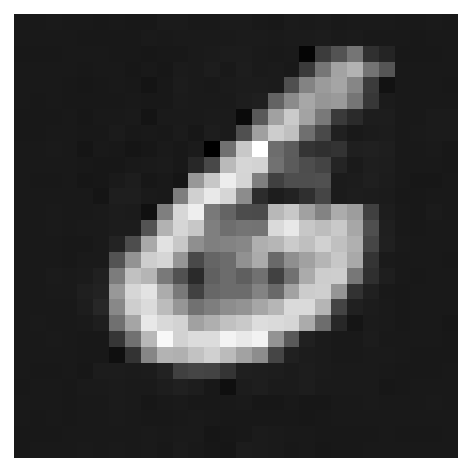}
& \includegraphics[trim={5mm 5mm 5mm 5mm},clip,width=\mywidth]{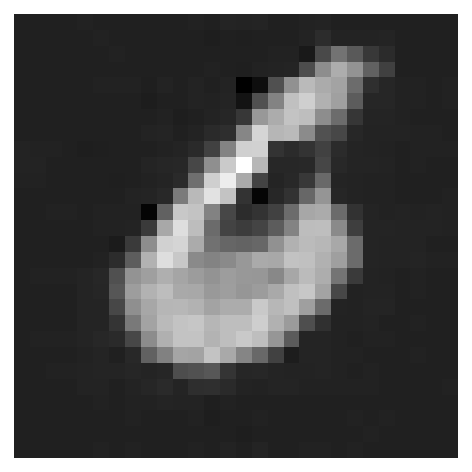}
&\includegraphics[trim={5mm 5mm 5mm 5mm},clip,width=\mywidth]{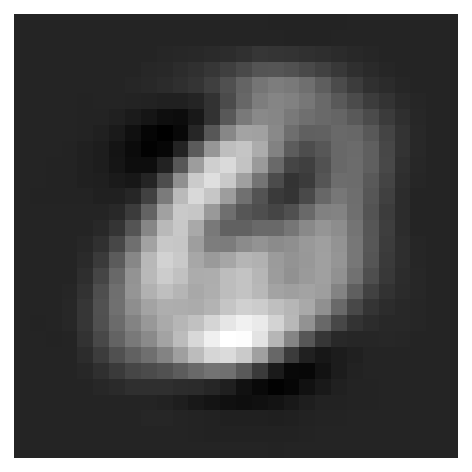}
& \includegraphics[trim={5mm 5mm 5mm 5mm},clip,width=\mywidth]{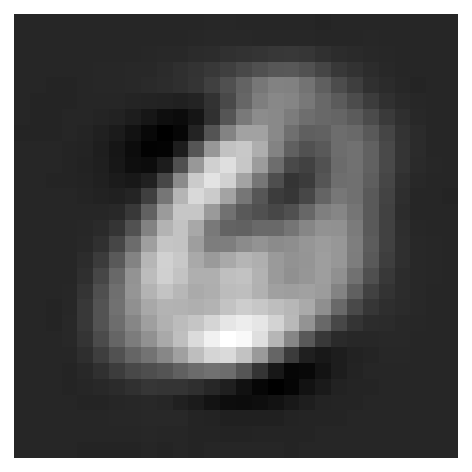}
&\includegraphics[trim={5mm 5mm 5mm 5mm},clip,width=\mywidth]{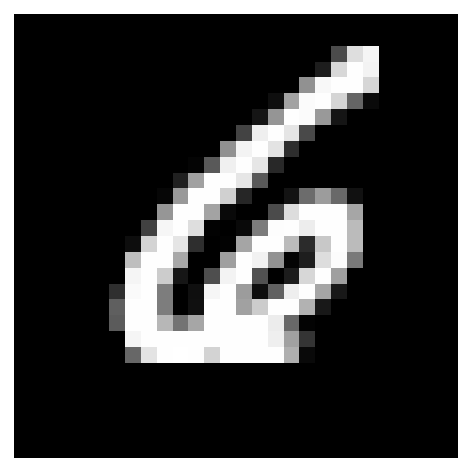}

 \\

\hline
\end{tabular}
\caption{Each line depicts the same MNIST image reconstructed by the four architectures. All autoencoders have the same latent dimension $d=8$. }\label{fig:mnist}
\end{figure}

Reconstruction metrics for $d\in\{1,2,4,8,16,32,64\}$ are reported in Table~\ref{tab:reconstruction_errors}; examples of reconstructions for $d=8$ are shown in Figure~\ref{fig:mnist}.

As with the paperclip test case, the reconstruction losses follow a consistent hierarchy across all values of $d$: the AE achieves the lowest error, followed by the Lenc-AE, while the Ldec-AE and LAE exhibit significantly higher and near-identical errors. This ranking reflects the expressivity constraints imposed by each architecture: the linear decoder shared by the Ldec-AE and LAE acts as a hard bottleneck that limits reconstruction quality regardless of the latent dimension, whereas the nonlinear decoder of the AE and Lenc-AE allows for much more accurate approximations. The example reconstructions at $d=8$ in Figure~\ref{fig:mnist} illustrate this clearly: the AE and Lenc-AE produce sharp, recognizable digits, while the Ldec-AE and LAE outputs remain blurry and poorly defined.

%% file: sections/results/08-office-video.tex
\subsubsection{Office Video}\label{sec:office}

This data set consists of 477 images of an office, captured by us while rotating the camera over more than 360 degrees, keeping it horizontal. The images theoretically form a closed 1-dimensional curve ($p=1$) in the image space, and can therefore in principle be learned by a nonlinear autoencoder with latent dimension $d=2$.

\begin{figure}[H]
\centering
\def\mywidth{24mm}
\begin{tabular}{ccc}
tSNE&AE without UMAP Loss&AE with UMAP Loss\\\hline
\includegraphics[trim={5mm 5mm 5mm 5mm},clip,width=\mywidth]{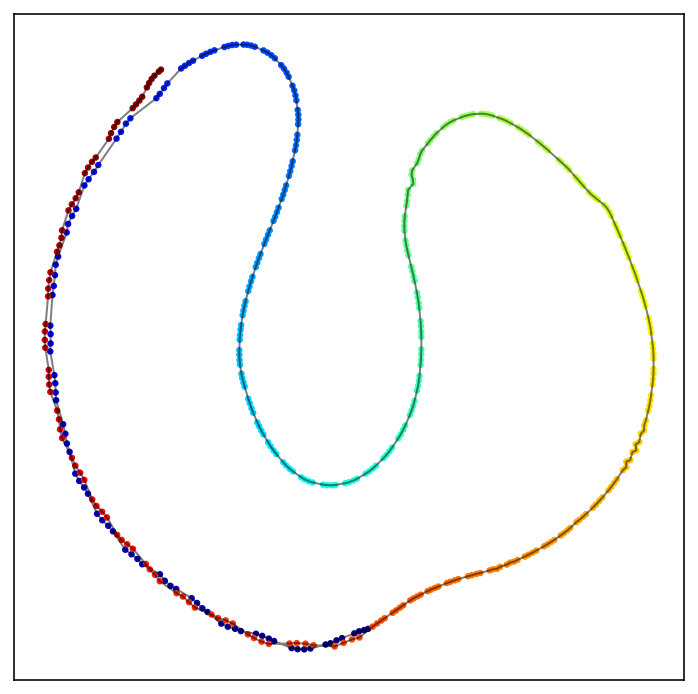} 
&\includegraphics[trim={5mm 5mm 5mm 5mm},clip,width=\mywidth]{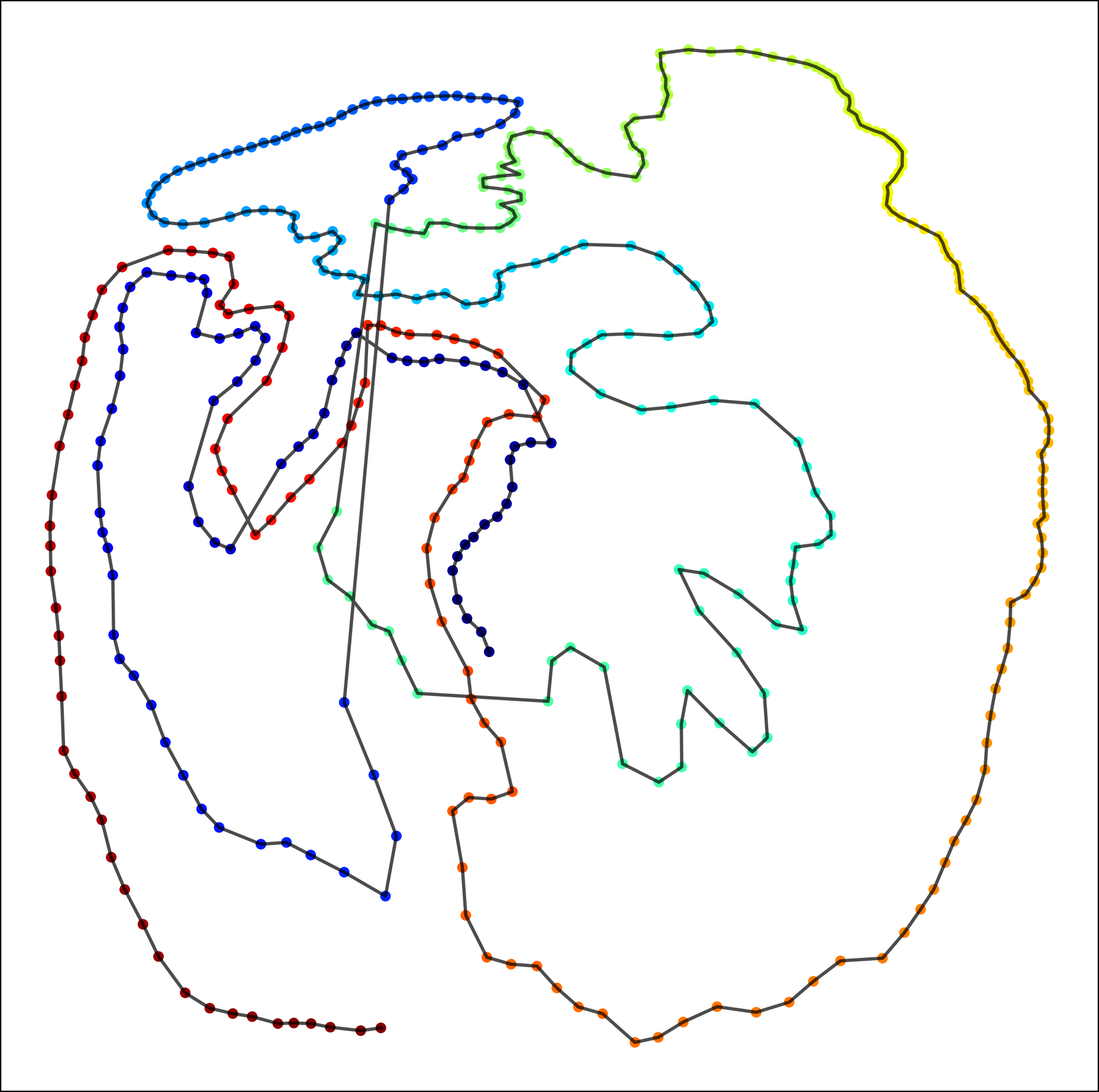} 
&\includegraphics[trim={5mm 5mm 5mm 5mm},clip,width=\mywidth]{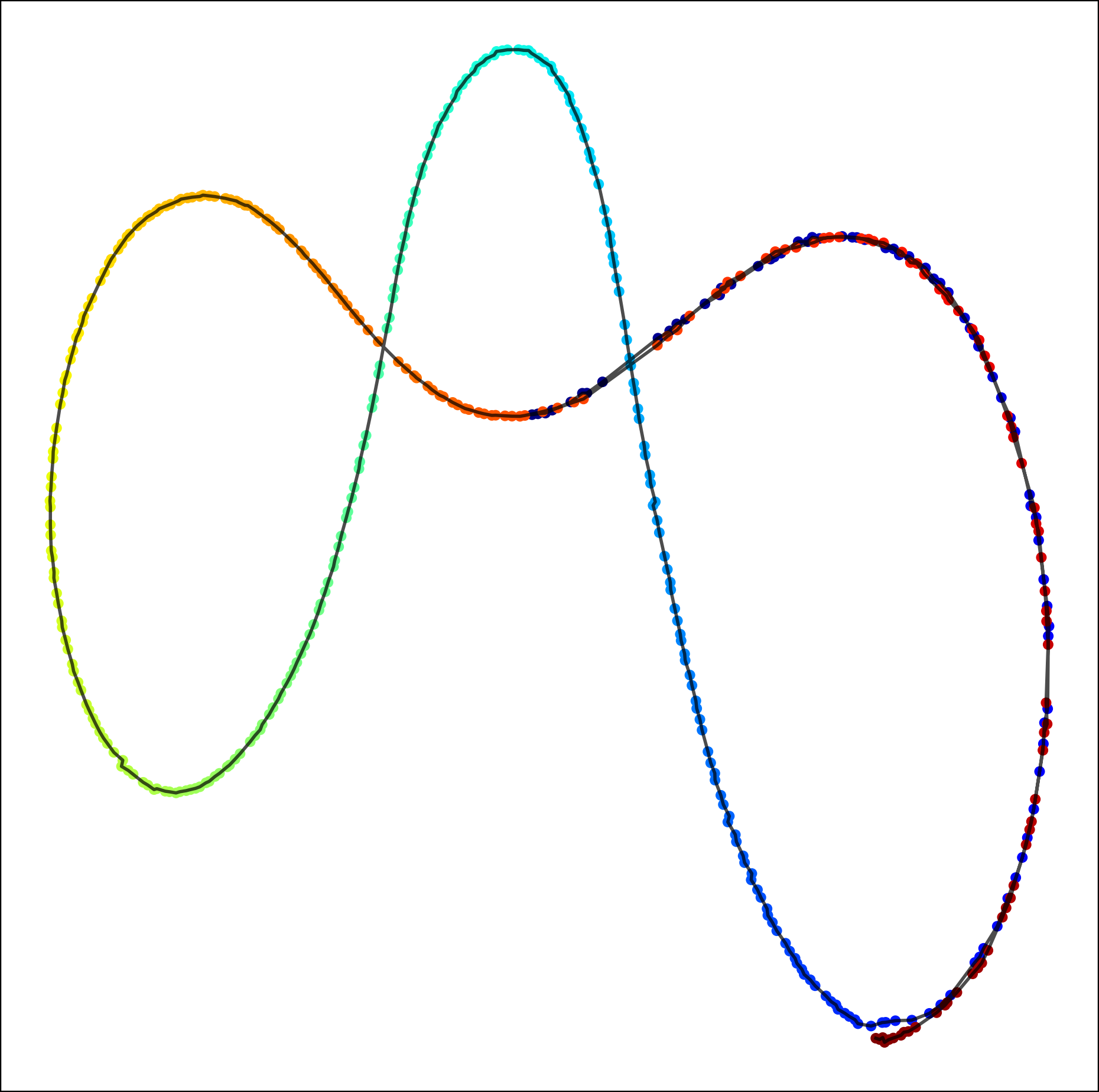} 
\end{tabular}
\caption{\label{fig:manifolds_video} Manifolds obtained from the different methods, embedded in $\mathbb{R}^2$.}
\end{figure}

Initial training runs of convolutional or fully-connected networks, using only a reconstruction loss on the images, revealed that preserving this manifold is challenging. When projecting all images into a 2D latent space, the resulting curve exhibits multiple self-intersections, whereas t-SNE applied to this data set successfully visualizes the circular topology (Figure~\ref{fig:manifolds_video}).

To encourage the autoencoder to produce a latent space that preserves the local structure of the data, we add a loss term inspired by UMAP, similarly to existing approaches~\citep{umapae,umapae2}. This loss pulls neighboring points in the latent space closer together while pushing away randomly sampled negative points.

Let $\mathbf{z}_i \in \mathbb{R}^d$ denote the latent representation of point $i$, and $\mathcal{N}(i)$ the set of its $k$ nearest neighbors. The latent similarity for each pair $(i,j)$ of neighbors is defined as:
\begin{equation}
q_{ij} = \frac{1}{1 + \|\mathbf{z}_i - \mathbf{z}_j\|^{2}},
\end{equation}

The positive loss, which pulls neighbors together, is:
\begin{equation}
\ell_{\mathrm{pos}} = - \frac{1}{N k} \sum_{i=1}^{N} \sum_{j \in \mathcal{N}(i)} \log \left( q_{ij} + \varepsilon \right),
\end{equation}
where $\varepsilon$ is a small term added for numerical stability.

The negative loss, which pushes these points apart, is:
\begin{equation}
\ell_{\mathrm{neg}} = - \frac{1}{N  n_{\mathrm{neg}}} \sum_{i=1}^{N} \sum_{j=1}^{ n_{\mathrm{neg}}} \log \left( 1 - \frac{1}{1 +  \, \|\mathbf{z}_i - \mathbf{z}_{ij}^- \|^{2}} + \varepsilon \right),
\end{equation}
where $n_{\mathrm{neg}}$ is the number of negative points $\mathbf{z}_{ij}^-$ per neighbor sampled randomly.

The total UMAP loss used during autoencoder training is then
\begin{equation}
\mathcal{L}_{\mathrm{UMAP}} = \ell_{\mathrm{pos}} + \ell_{\mathrm{neg}}.
\end{equation}

Adding this loss during training yields latent representations with fewer self-intersections (though a few may remain for $d=2$). For $d=3$, the manifold is consistently well reconstructed for all architectures, but linear decoders are generally insufficient to reconstruct high-quality images.

\begin{figure}[h!]
\centering
\def\mywidth{24mm}
\begin{tabular}{m{2cm}m{2cm}m{2cm}m{2cm}m{2cm}m{2.cm}}
&AE&Lenc-AE&Ldec-AE&LAE&Objective\\\hline
$d=2$ &
\includegraphics[trim={5mm 5mm 5mm 5mm},clip,width=\mywidth]{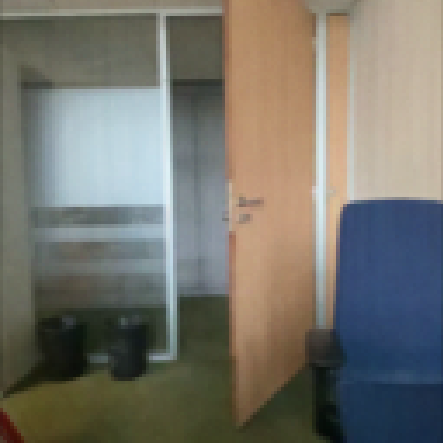} 
&\includegraphics[trim={5mm 5mm 5mm 5mm},clip,width=\mywidth]{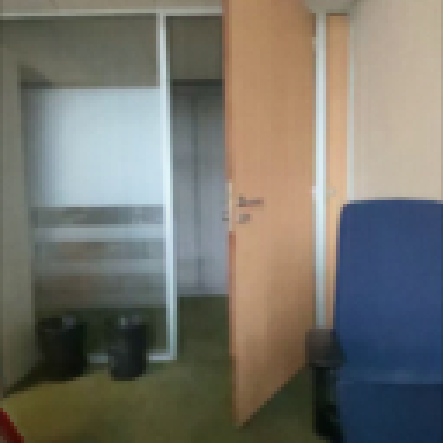} 
&\includegraphics[trim={5mm 5mm 5mm 5mm},clip,width=\mywidth]{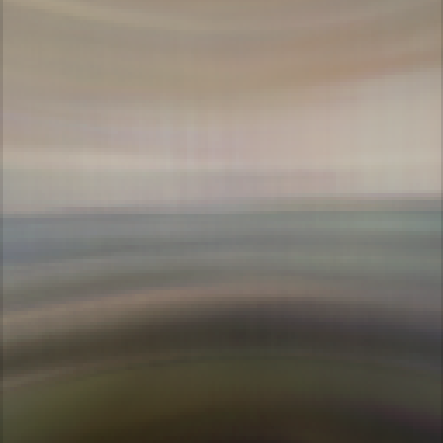} 
&\includegraphics[trim={5mm 5mm 5mm 5mm},clip,width=\mywidth]{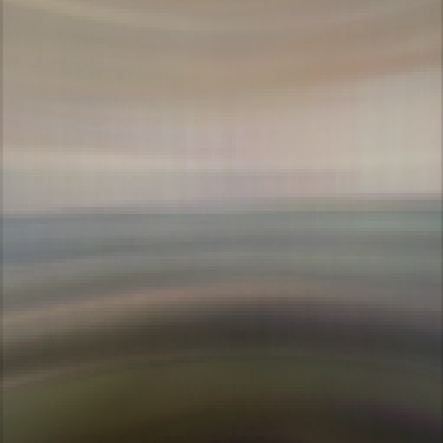} 
& \includegraphics[trim={5mm 5mm 5mm 5mm},clip,width=\mywidth]{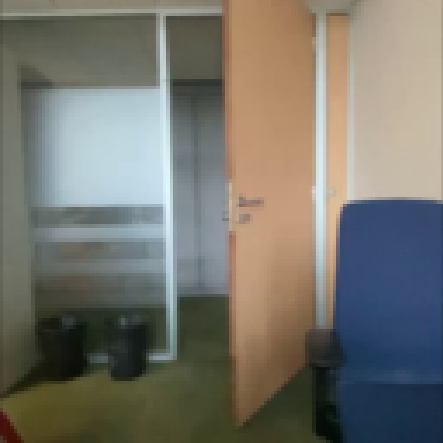} \\
&\includegraphics[trim={5mm 5mm 5mm 5mm},clip,width=\mywidth]{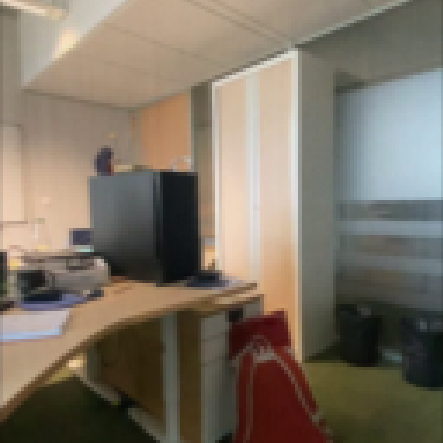} 
&\includegraphics[trim={5mm 5mm 5mm 5mm},clip,width=\mywidth]{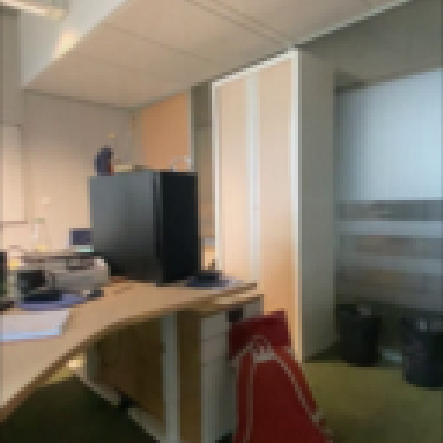} 
&\includegraphics[trim={5mm 5mm 5mm 5mm},clip,width=\mywidth]{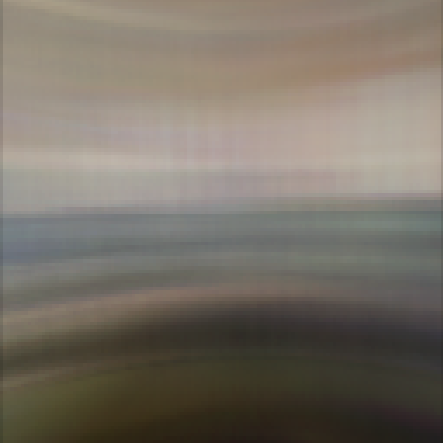} 
&\includegraphics[trim={5mm 5mm 5mm 5mm},clip,width=\mywidth]{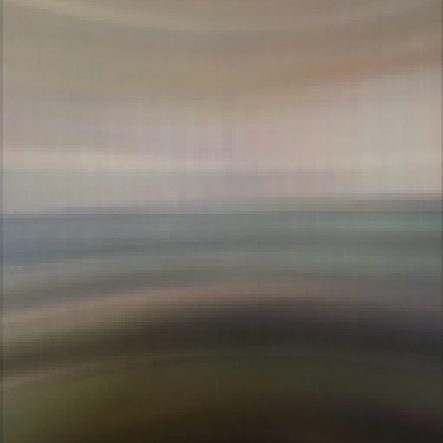} 
& \includegraphics[trim={5mm 5mm 5mm 5mm},clip,width=\mywidth]{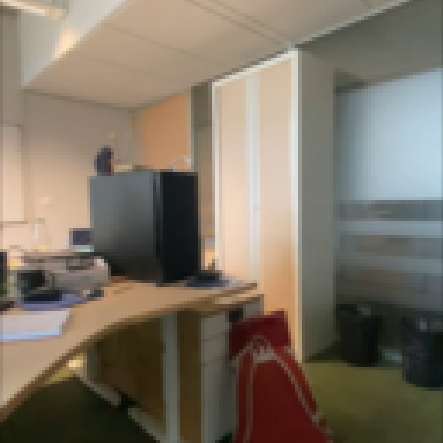} \\
latent space &
\includegraphics[trim={5mm 5mm 5mm 5mm},clip,width=\mywidth]{video2/aedim2lat.png}
&\includegraphics[trim={5mm 5mm 5mm 5mm},clip,width=\mywidth]{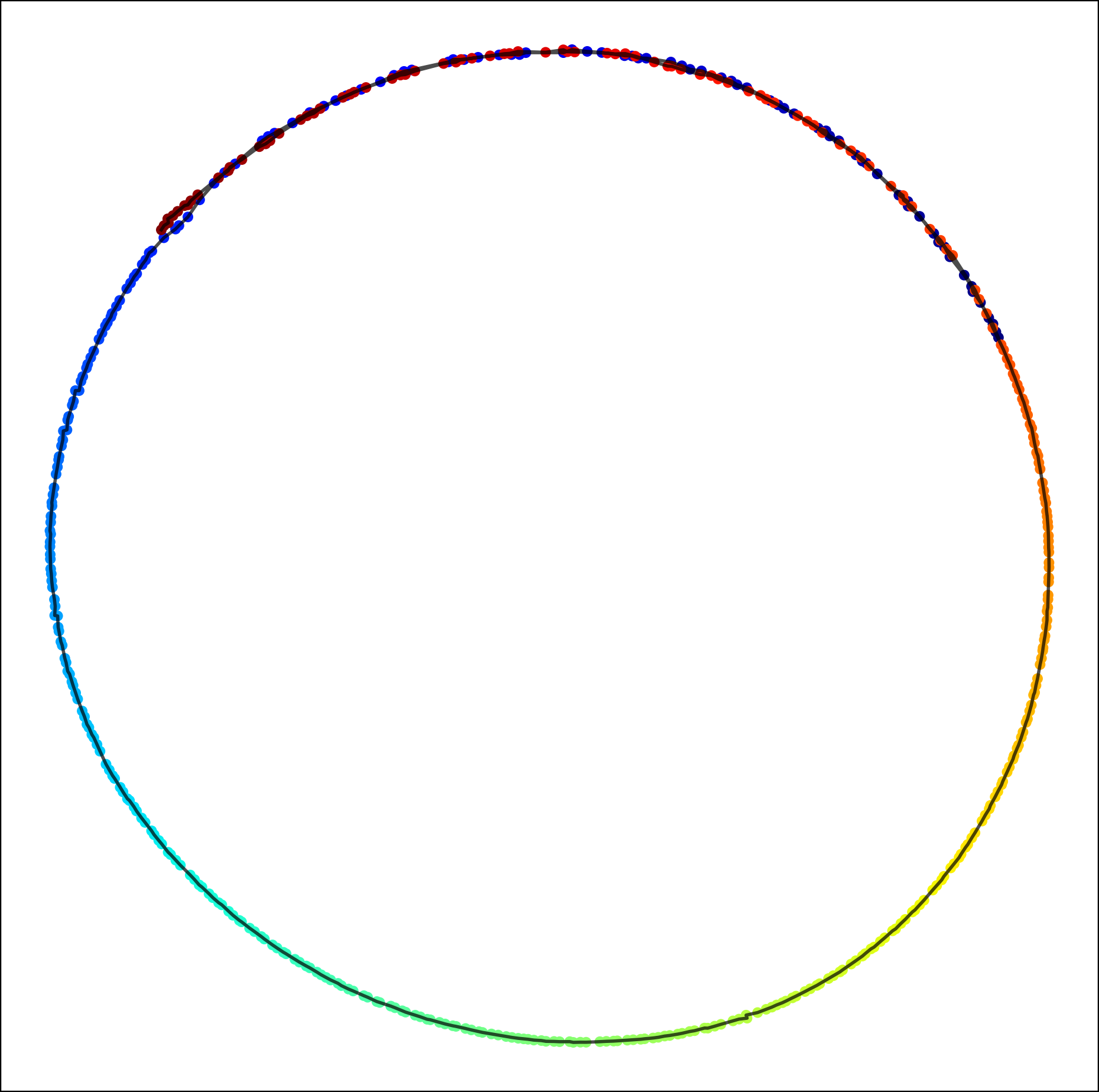} 
& \includegraphics[trim={5mm 5mm 5mm 5mm},clip,width=\mywidth]{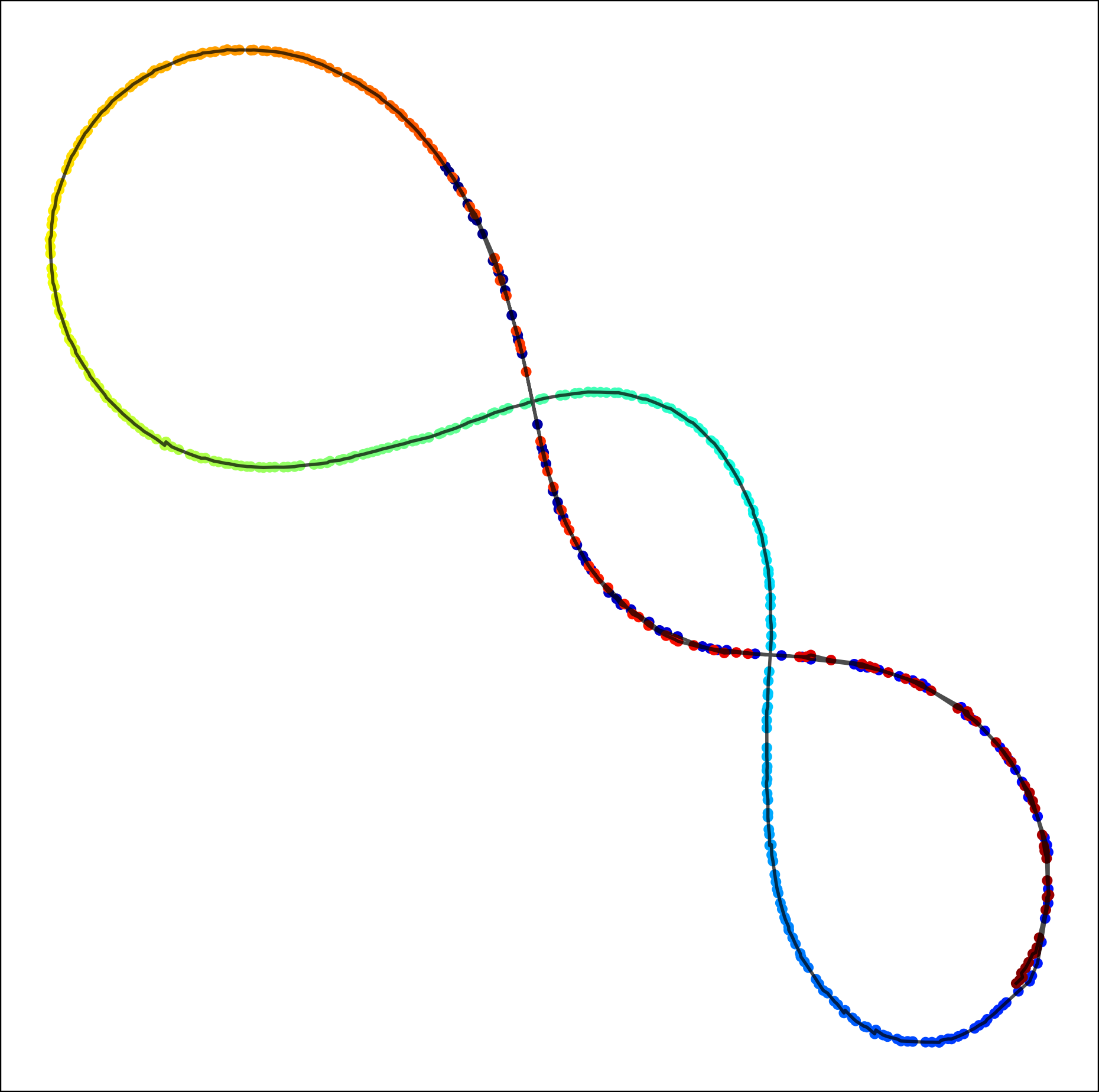} 
&\includegraphics[trim={5mm 5mm 5mm 5mm},clip,width=\mywidth]{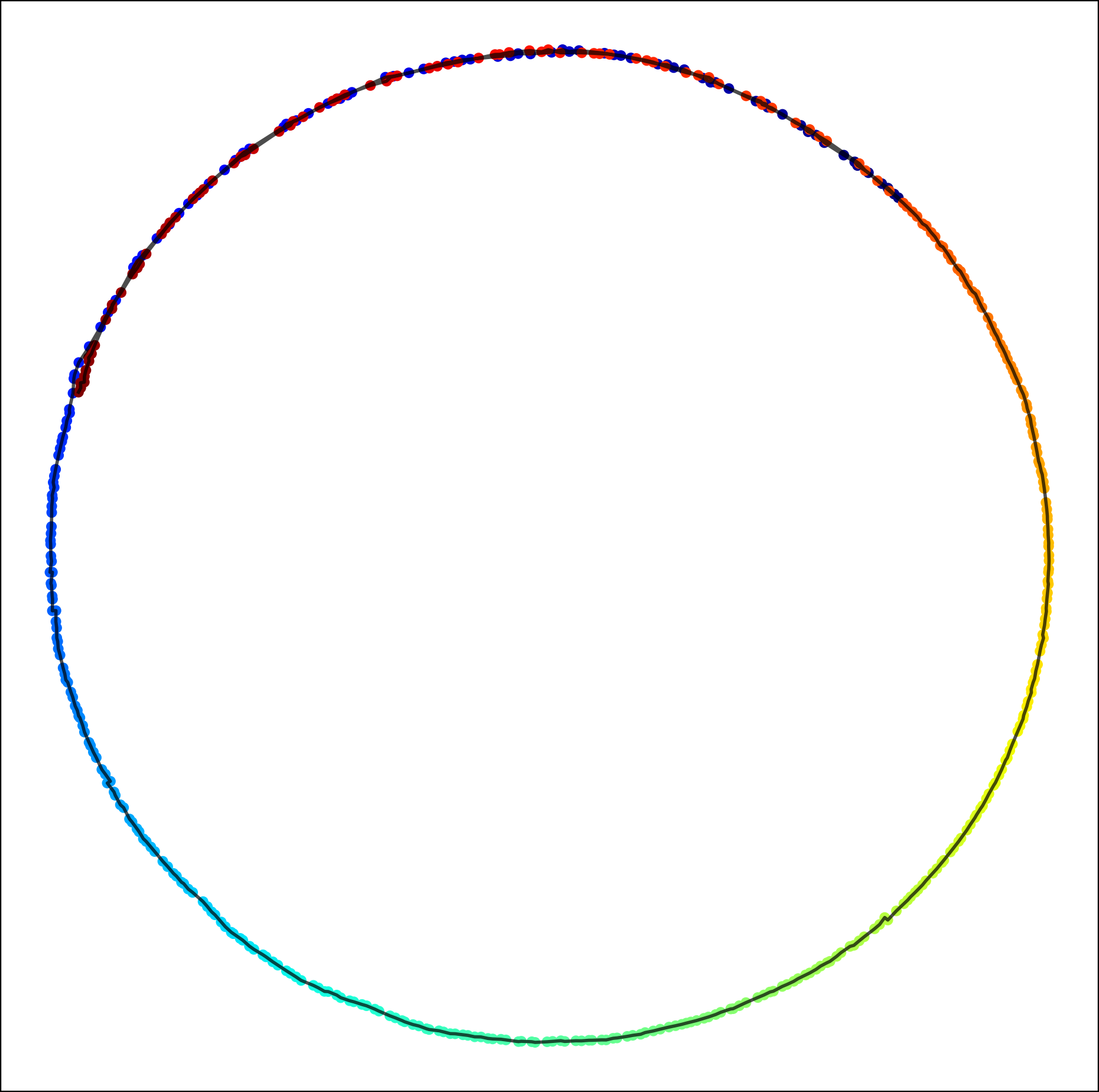} 
& \\
\hline

$d=3$ &
\includegraphics[trim={5mm 5mm 5mm 5mm},clip,width=\mywidth]{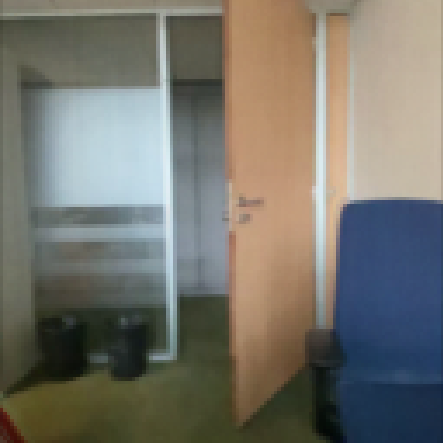} 
&\includegraphics[trim={5mm 5mm 5mm 5mm},clip,width=\mywidth]{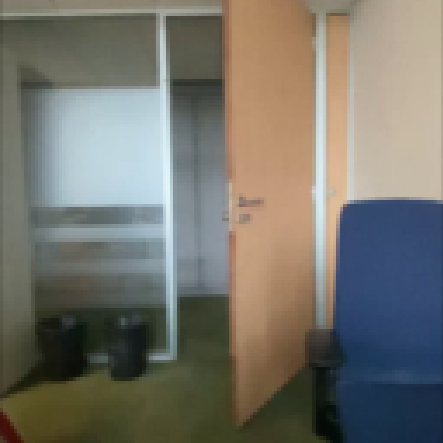} 
&\includegraphics[trim={5mm 5mm 5mm 5mm},clip,width=\mywidth]{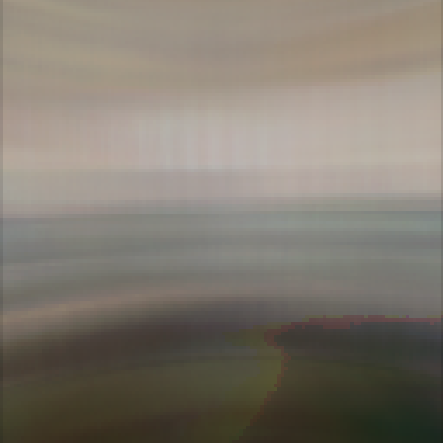} 
&\includegraphics[trim={5mm 5mm 5mm 5mm},clip,width=\mywidth]{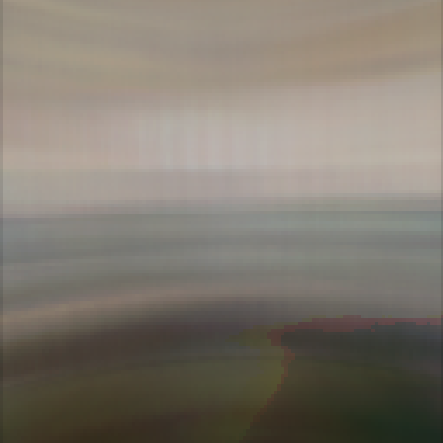} 
& \includegraphics[trim={5mm 5mm 5mm 5mm},clip,width=\mywidth]{video/img1obj.png} \\
&\includegraphics[trim={5mm 5mm 5mm 5mm},clip,width=\mywidth]{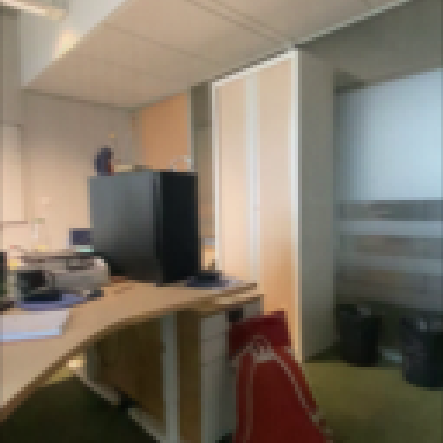} 
&\includegraphics[trim={5mm 5mm 5mm 5mm},clip,width=\mywidth]{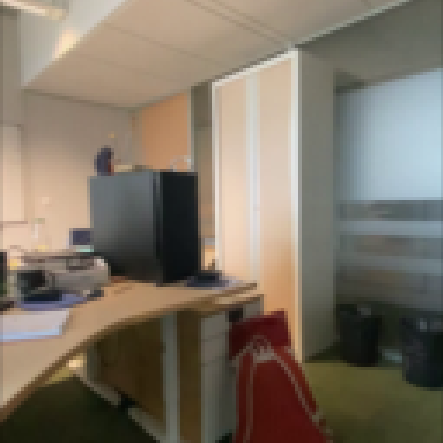} 
&\includegraphics[trim={5mm 5mm 5mm 5mm},clip,width=\mywidth]{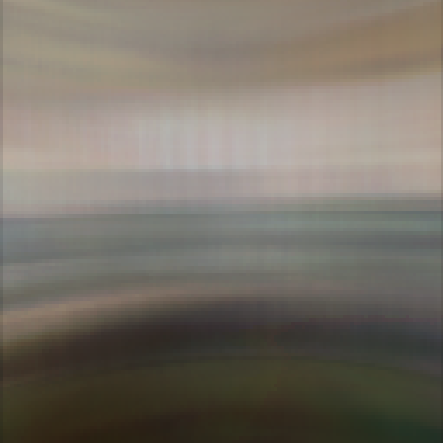} 
&\includegraphics[trim={5mm 5mm 5mm 5mm},clip,width=\mywidth]{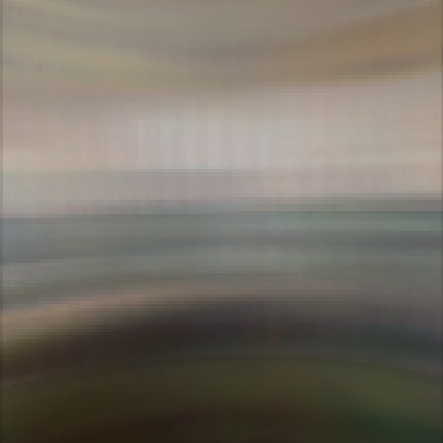} 
& \includegraphics[trim={5mm 5mm 5mm 5mm},clip,width=\mywidth]{video/img2obj.png} \\
latent space &
\includegraphics[trim={5mm 5mm 5mm 5mm},clip,width=\mywidth]{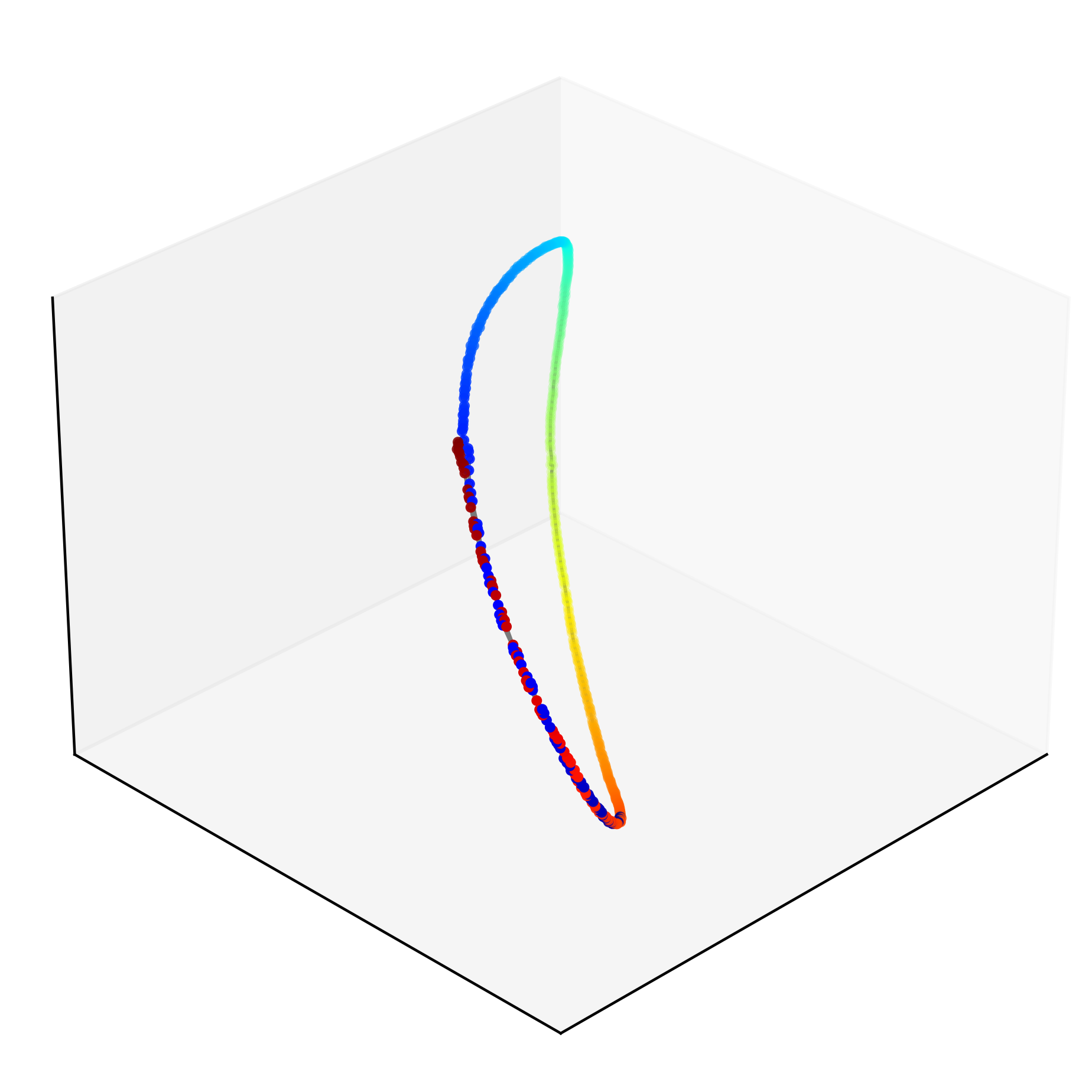}
&\includegraphics[trim={5mm 5mm 5mm 5mm},clip,width=\mywidth]{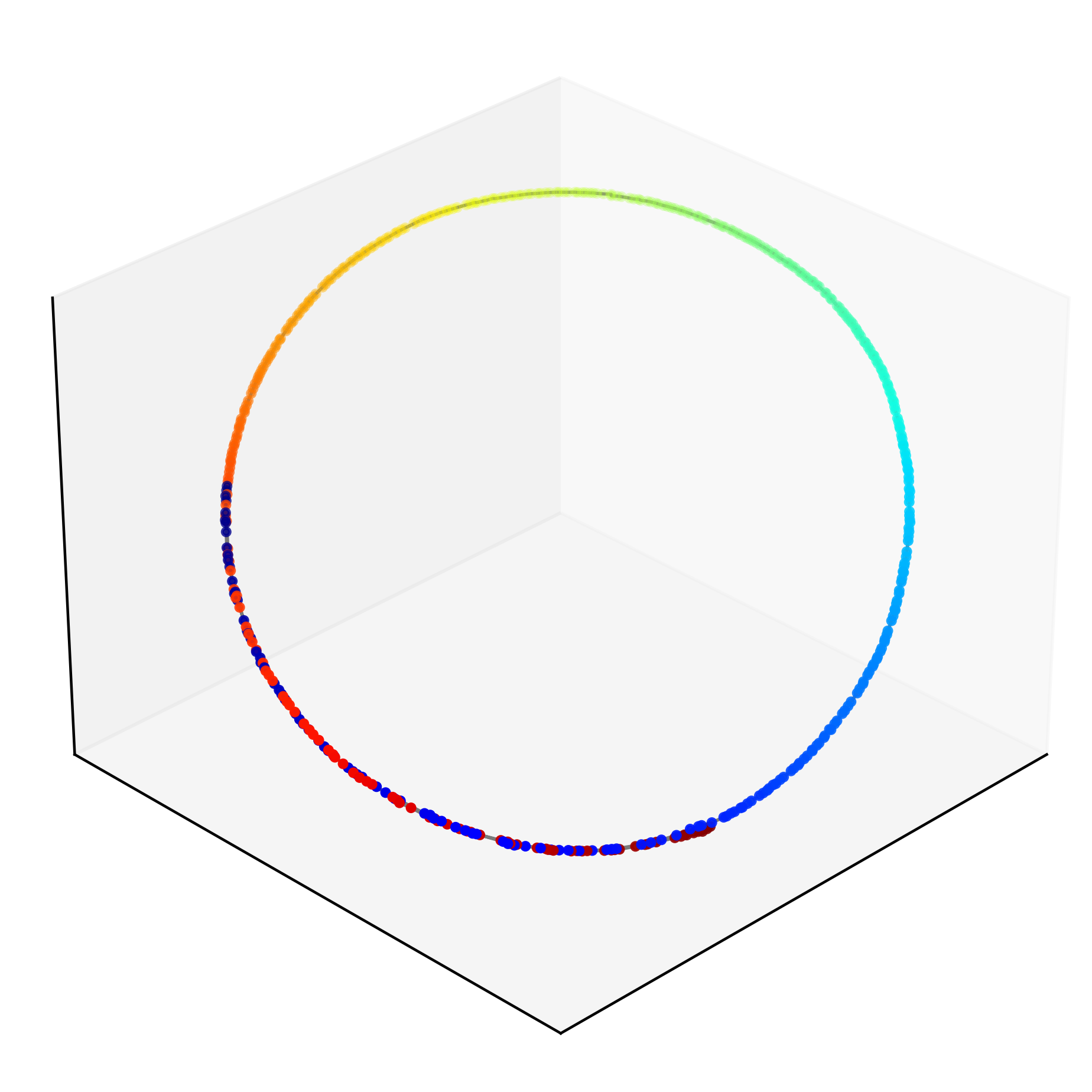} 
& \includegraphics[trim={5mm 5mm 5mm 5mm},clip,width=\mywidth]{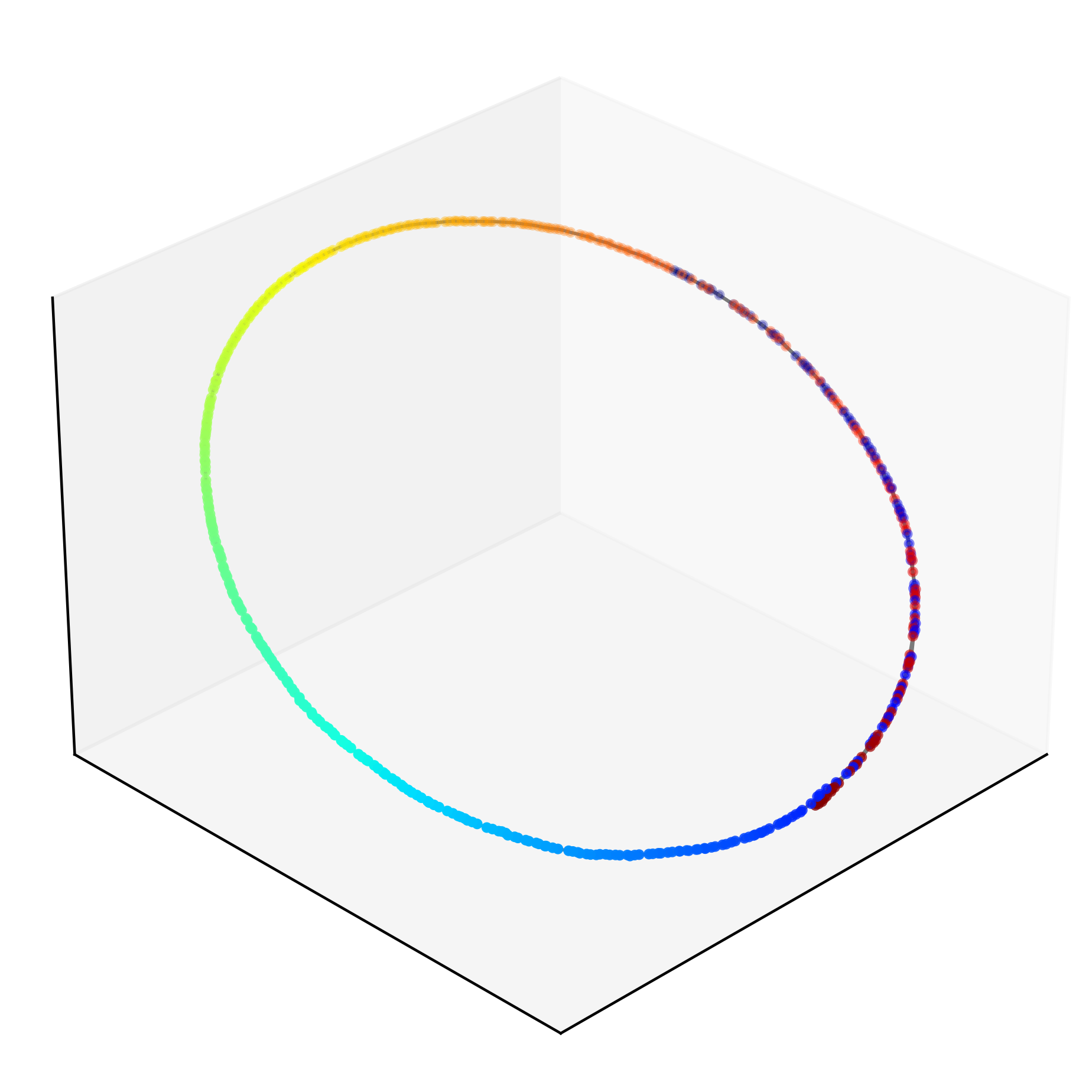} 
&\includegraphics[trim={5mm 5mm 5mm 5mm},clip,width=\mywidth]{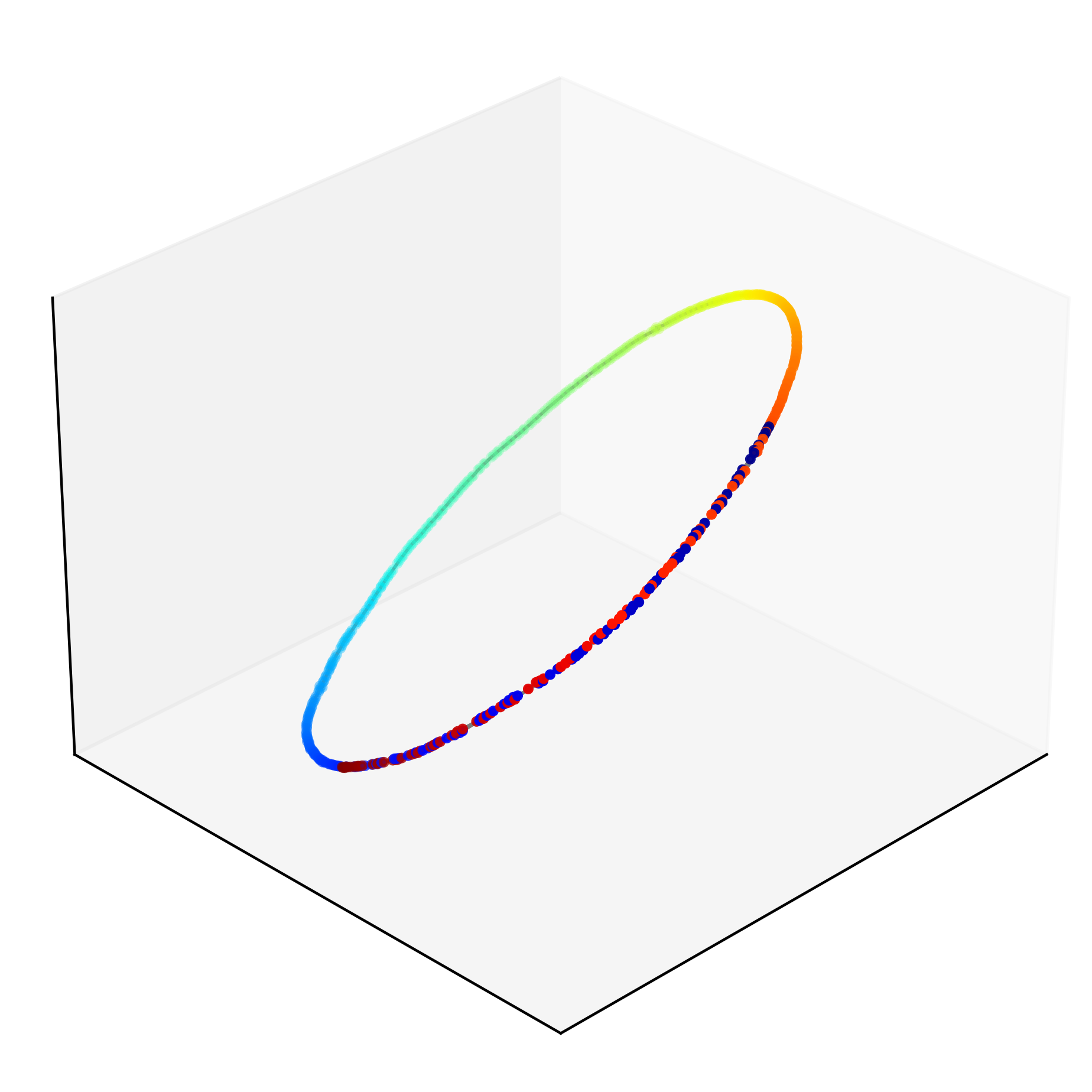} 
& \\
\hline
\end{tabular}
\caption{Each line depicts the same Office image reconstructed by the four architectures for $d\in\{2,3\}$. The latent space visualization is also provided; each dot represents an image. The circular topology is clearly visible, although for $d=2$ the encoder sometimes fails to learn a projection without overlap.} \label{fig:video} 
\end{figure}

The results are shown in Figures~\ref{fig:manifolds_video} and~\ref{fig:video}. Without the UMAP loss, Lenc-AE slightly outperforms the standard AE in terms of reconstruction error (see Table~\ref{tab:reconstruction_errors}). When the UMAP loss is included, the AE performs slightly better than the Lenc-AE. Interestingly, although the UMAP loss affects only the encoder weights, one might have expected it to perform poorly with a linear encoder due to its limited flexibility. Yet this is not the case: a linear encoder operating on the high-dimensional image space retains enough degrees of freedom to learn a smooth latent manifold.

%% file: sections/05-discussion.tex
\section{Synthesis and Discussion}
\label{sec:discussion}

The reconstruction errors for all test cases are provided in Table~\ref{tab:reconstruction_errors}. 
These results indicate a consistent hierarchy across all test cases: the AE and Lenc-AE perform comparably, with the AE slightly ahead in most cases—though not universally, as the Lenc-AE occasionally matches or surpasses it. The Ldec-AE and LAE remain close to each other throughout, but consistently lag behind the AE and Lenc-AE in both reconstruction accuracy and manifold unfolding quality. This confirms that the nonlinear decoder is the dominant factor in representational capacity, while the linearity of the encoder in Lenc-AE incurs only a marginal cost.

\begin{table}[h!]
\centering
\begin{tabular}{llrrrr}
\toprule
Data set & $d$ & AE & Lenc-AE & Ldec-AE & LAE \\
\midrule
\multirow{2}{*}{Helix}
 & 1 & 49.88 & \textbf{0.341} & 1000 & 999.95 \\
 & 2 & \textbf{5.57} & 11.56 & 422.60 & 422.56 \\
\midrule
\multirow{2}{*}{Knot}
 & 1 & \textbf{112.17} & 595.54 & 999.93 & 1000 \\
 & 2 & \textbf{4.60} & 4.74 & 445.13 & 445.13 \\
\midrule
S-curve & 2 & \textbf{10.19} & 100.75 & 999.92 & 1000 \\
\midrule
Swiss roll & 2 & \textbf{113.61} & 206.27 & 1000 & 990.72 \\
\midrule
\multirow{3}{*}{Beam}
 & 1 & \textbf{12.46} & 81.06 & 1000 & 999.88 \\
 & 2 & 14.88 & \textbf{2.56} & 26.46 & 26.42 \\
 & 3 & \textbf{0.170} & 0.199 & 5.71 & 5.63 \\
\midrule
\multirow{7}{*}{Paperclip}
 & 1  & 861.64 & \textbf{853.71} & 995.08 & 1000 \\
 & 2  & \textbf{312.27} & 314.31 & 501.04 & 499.73 \\
 & 4  & \textbf{88.79} & 118.34 & 216.02 & 214.13 \\
 & 8  & 16.13 & \textbf{11.97} & 71.26 & 70.57 \\
 & 16 & \textbf{1.81} & 2.67 & 3.00 & 7.51 \\
 & 32 & \textbf{1.73} & 1.75 & 2.37 &3.85 \\
 & 64 & \textbf{0.92} & 1.86 & 2.82 & 3.19 \\
\midrule
\multirow{7}{*}{MNIST}
 & 1  & \textbf{861.39} & 900.77 & 1000 & 999.81 \\
 & 2  & \textbf{735.94} & 773.76 & 913.67 & 913.51 \\
 & 4  & \textbf{551.87} & 602.92 & 772.23 & 772.09 \\
 & 8  & \textbf{339.75} & 406.84 & 602.38 & 602.06 \\
 & 16 & \textbf{183.41} & 231.63 & 435.31 & 434.03 \\
 & 32 & \textbf{99.28} & 121.65 & 276.20 & 273.35 \\
 & 64 & \textbf{50.23} & 50.56 & 156.74 & 147.53 \\
\midrule
\multirow{3}{*}{Office}
 & 1 & \textbf{360.70} & 366.56 & 1000 & 982.40 \\
 & 2 & 73.31 & \textbf{55.72} & 809.38 & 806.45 \\
 & 3 & 52.78 & \textbf{41.06} & 750.73 & 733.13 \\
\cline{2-6} %
\multirow{2}{*}{Office (w/ UMAP)}
 & 2 & \textbf{40.76} & 48.68 & 821.11 & 815.24 \\
 & 3 & \textbf{29.31} & 36.51 & 797.65 & 794.72 \\
\bottomrule
\end{tabular}
\caption{Reconstruction error normalized with respect to the worst result for each test case. Best results in bold.}\label{tab:reconstruction_errors}
\end{table}

The results presented above admit a natural geometric interpretation that clarifies when and why Lenc-AE matches the performance of a fully nonlinear AE. A linear encoder $\enc : \mathbb{R}^D \rightarrow \mathbb{R}^d$ is, by definition, a projection onto a $d$-dimensional linear subspace. For such an encoder to serve as a valid coordinate system for the data manifold $\mathcal{M}$, it is necessary that this projection be injective when restricted to $\mathcal{M}$—that is, distinct points on $\mathcal{M}$ must map to distinct latent coordinates. When such an injective linear projection exists at the minimal embedding dimension $d$, a sufficiently expressive nonlinear decoder can, in principle, perfectly reconstruct $\mathcal{M}$ from its linear projection, and Lenc-AE is therefore as expressive as a fully nonlinear AE.

This condition is satisfied for the Helix (Figure~\ref{fig:plot_helix}, $d=1$), where the projection onto the central axis is injective, and for the Beam data set (Figure~\ref{fig:plot_beam}, $d=2$), where the two-dimensional load space admits a linear parametrization of the displacement field manifold. It is also satisfied for the Office data set (Figure~\ref{fig:video}, $d=2$), where despite the circular topology of the manifold—which prevents perfect reconstruction at $d=p=1$ for any autoencoder—an injective linear projection onto $\mathbb{R}^2$ exists, and Lenc-AE is able to learn one, indeed matching AE in this setting. Conversely, the condition fails for manifolds whose rolled or folded geometry inevitably produces overlaps under any linear projection, such as the Swiss roll (Figure~\ref{fig:swissroll}) and the S-curve (Figure~\ref{fig:Scurve}). For higher-dimensional real-world data sets such as MNIST and the Paperclip (Table~\ref{tab:reconstruction_errors}), the condition is harder to verify analytically, but the competitive performance of Lenc-AE suggests that approximate injectivity is achievable in practice.

This geometric picture also clarifies the asymmetry between Lenc-AE and Ldec-AE. When imposing linearity on the decoder, perfect reconstruction in $d$ dimensions is only possible if the data manifold is itself a linear subspace of $\mathbb{R}^D$—a far more restrictive condition. By contrast, a linear encoder with a nonlinear decoder requires only that an injective linear projection of $\mathcal{M}$ onto $\mathbb{R}^d$ exists, which is compatible with curved or deformed structures. This explains why Ldec-AE suffers more from underfitting than Lenc-AE across all test cases: a nonlinear encoder alone cannot compensate for a decoder that is too rigid to reconstruct the manifold's geometry.

More broadly, these findings confirm that the nonlinear decoder is the critical component for manifold learning in autoencoders. This separation of roles—a linear encoder providing an interpretable, analytically tractable latent representation, and a nonlinear decoder handling the geometric complexity of reconstruction—offers a compelling compromise between expressivity and interpretability. Lenc-AE may therefore serve as a computationally efficient and analytically transparent alternative to standard AEs in settings where the injectivity condition is met, or approximately met.

Finally, it is worth noting that even when the injectivity condition fails at $d = p$, satisfactory reconstruction may still be achieved in the overcomplete regime $d > p$. The additional latent dimensions provide the nonlinear decoder with sufficient degrees of freedom to resolve the ambiguities introduced by a non-injective linear projection, recovering high reconstruction quality despite the encoder constraint. This is consistent with the results observed for the Paperclip data set (Figure~\ref{tab:paperclip}) and for MNIST at large $d$ (Table~\ref{tab:reconstruction_errors}).

%% file: sections/06-conclusion.tex
\section{Conclusion and Further Work}

This work provides a systematic comparison of four autoencoder architectures (AE, Lenc-AE, Ldec-AE, and LAE) across a range of synthetic, mechanical, and image data sets. The central finding is that a linear encoder, when paired with a nonlinear decoder, preserves most of the representational capacity of a fully nonlinear autoencoder, while offering complementary advantages in interpretability, computational efficiency, and latent space regularity. The key geometric condition underpinning this result is the existence of an injective linear projection of the data manifold onto the latent space: when this condition holds—exactly or approximately—Lenc-AE matches fully nonlinear AE in reconstruction quality. Conversely, linearizing the decoder imposes a far more restrictive constraint on the data geometry, explaining the consistently poor performance of Ldec-AE and LAE across all test cases.

By clarifying the representational power of partially linear autoencoder architectures, this work opens the door to several domain-specific developments that can exploit the structural properties of Lenc-AE and Ldec-AE.
A particularly promising direction concerns model-order reduction in structural mechanics. Classical substructuring methods such as Craig–Bampton reduction~\citep{craigbrampton} and Guyan reduction~\citep{guyan} rely on fully linear encoder-decoder pairs: interface and boundary degrees of freedom are encoded via physics-informed linear transformations, and internal states are reconstructed by linear superposition of precomputed mode shapes. This linearity is not merely a simplification but a structural requirement, ensuring compatibility at substructure interfaces and enabling modular assembly of reduced-order models. These methods therefore correspond precisely to the LAE architecture, and inherit its fundamental limitation: they cannot capture nonlinear deformation regimes, such as those arising under large displacements or material nonlinearities.
The Lenc-AE architecture offers a natural generalization of this framework. Retaining a linear encoder preserves the interface compatibility conditions that make substructuring methods practically viable, while replacing the linear decoder with a nonlinear neural network enables the reduced model to represent complex, nonlinear deformation modes. This combination—structurally principled at the encoding stage, expressively flexible at the decoding stage—is precisely the regime in which Lenc-AE has been shown to perform competitively with fully nonlinear AE. 

Developing a Craig–Bampton-style reduction method based on the Lenc-AE architecture, with theoretical guarantees on interface compatibility and numerical studies on nonlinear structural benchmarks, constitutes a concrete and practically motivated avenue for future research.
More broadly, the injectivity condition identified in Section~\ref{sec:discussion} suggests a principled criterion for choosing between AE and Lenc-AE in a given application: whenever the solution manifold admits an approximately injective linear projection—a condition that is naturally satisfied in physical simulation settings where the solution space inherits regularity from the underlying PDE—Lenc-AE provides a transparent and computationally efficient alternative to fully nonlinear approaches. Characterizing this condition more precisely for specific classes of mechanical problems, and developing practical tests for its approximate validity, represents a further line of inquiry.